\documentclass[9.5pt,cspaper,compsoc,final]{IEEEtran}
\usepackage{packages}
\usepackage{customized}
\usepackage{ragged2e}

\begin{document}

\title{Fisheye Distortion Rectification from\\ Deep Straight Lines}
\author{Zhu-Cun~Xue, \and Nan~Xue,
\and Gui-Song~Xia%,~\IEEEmembership{Senior Member,~IEEE}
\thanks{All the authors are with Department of Computer Science and State Key Lab. LIESMARS, Wuhan University, Wuhan, 430072, China. E-mail: \{{\em zhucun.xue, xuenan, guisong.xia}\}@whu.edu.cn}
}
\IEEEtitleabstractindextext{
\justify
\begin{abstract}
	This paper presents a novel line-aware rectification network (LaRecNet) to address the problem of fisheye distortion rectification based on the classical observation that \emph{straight lines in 3D space should be still straight in image planes}. Specifically, the proposed LaRecNet contains three sequential modules to (1) learn the distorted straight lines from fisheye images; (2) estimate the distortion parameters from the learned heatmaps and the image appearance; and (3) rectify the input images via a proposed differentiable rectification layer. To better train and evaluate the proposed model, we create a synthetic line-rich fisheye (SLF) dataset that contains the distortion parameters and well-annotated distorted straight lines of fisheye images. 
	The proposed method enables us to simultaneously calibrate the geometric distortion parameters and rectify fisheye images.
	Extensive experiments demonstrate that our model achieves state-of-the-art performance in terms of both geometric accuracy and image quality on several evaluation metrics.
	In particular, the images rectified by LaRecNet achieve an average reprojection error of 0.33 pixels on the SLF dataset and produce the highest peak signal-to-noise ratio (PSNR) and  structure similarity index (SSIM) compared with the groundtruth.
\end{abstract}
}

\maketitle

\IEEEraisesectionheading{
\section{Introduction}\label{sec:introduction}
}

Fisheye cameras enable us to capture images with an ultrawide field of view (FoV) and have the potential to benefit many machine vision tasks, such as structure from motion (SfM), simultaneous localization and mapping (SLAM) and autonomous driving, by perceiving a well-conditioned scene coverage with fewer images than would otherwise be  required. However, the ultrawide FoV characteristics of fisheye cameras are usually achieved by nonlinear mapping functions~\cite{weng1992camera, ricolfe2010lens, yang2005nonlinear}, which always lead to severe geometric distortions to the sensed images and harm the subsequent vision computing process, e.g., 3D reconstructions and recognition. Therefore, when developing vision systems equipped with fisheye cameras~\cite{bertozzi2000vision,xiong1997creating,huang20176,szeliski1997creating}, it is often a prerequisite to estimate the camera parameters and rectify the geometric distortions.

In this paper, we address the problem of rectifying geometric distortions in fisheye images, which is commonly studied as a coupled problem of camera calibration~\cite{zhang2000flexible,Grossberg2001A,Sturm2004A,kannala2006generic,Scaramuzza2006A,heikkila2000geometric}.
Given a specified model of image formation (\eg, the perspective camera model), the distortion parameters are regarded as parts of the camera intrinsics and then estimated from the 2D-3D correspondences established by using certain predefined calibration patterns (\eg, cubes and planar checkerboards).
The geometric distortions in images are subsequently rectified with the estimated camera intrinsics. However, the methods of this category have to face the problem of establishing the 2D-3D correspondences, which is the vital factor for the calibration/rectification accuracy and often requires laborious computations~\cite{zhang2000flexible,Grossberg2001A}. 

To overcome the abovementioned drawbacks, self-calibration methods~\cite{stein1997lens,Faugeras1992Camera,Maybank1992A,devernay2001straight, Range, barreto2009automatic,melo2013unsupervised,bukhari2013automatic,aleman2014automatic,zhang2015line} have addressed the problem based on geometric invariant features (\eg, straight lines, vanishing points, conics) of images instead of using known 3D information of scenes.
Studies of these approaches have reported promising performances on fisheye distortion rectification when the specified geometric features in fisheye images can be accurately detected. Nevertheless, it is worth noticing that the involved detection of the geometric features in fisheye images itself is another challenging problem in computer vision.

\begin{figure}[t!]
	\centering
	\includegraphics[width=0.99\linewidth]{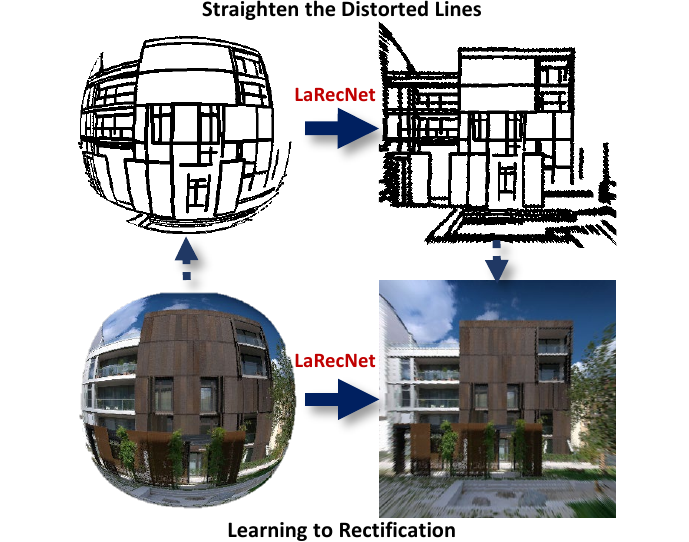}
\vspace{-2mm}
\caption{Geometric constraint in fisheye images: the distorted lines generated by fisheye projection should be straight in normal perspective images.	This paper investigates how to efficiently embed this geometric constraint into deep models for fisheye image rectification.}
\label{fig:map}
\vspace{-3mm}
\end{figure}

\begin{figure*}[t!]
	\centering
	\includegraphics[width=0.87\linewidth]{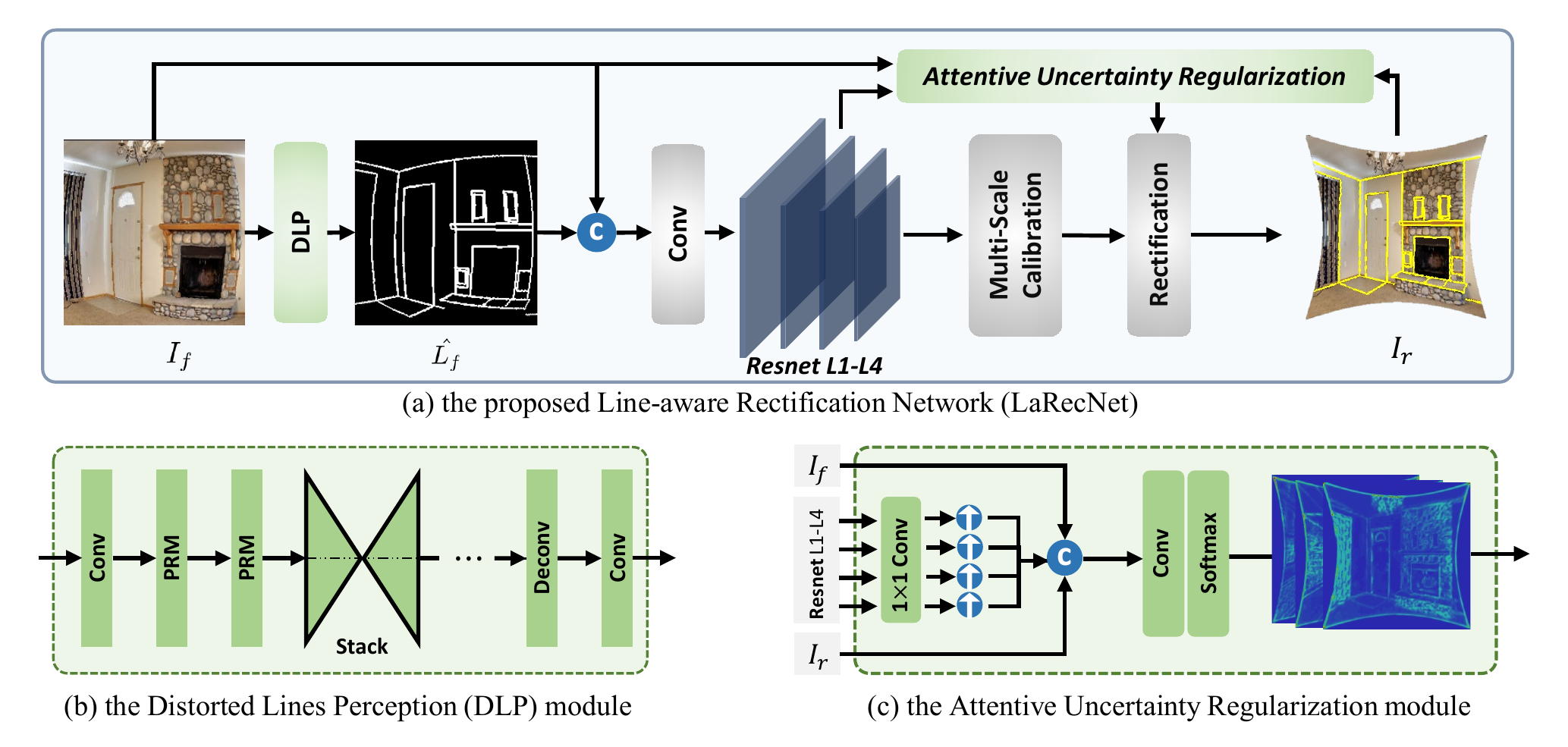}
	\vspace{-3mm}
\caption{Overview of our line-aware rectification network (LaRecNet). LaRecNet estimates the distortion parameters by learning how to straighten distorted lines; it mainly contains the distorted line perception (DLP) module, a multiscale calibration module, a differentiable rectification module, and an attentive uncertainty regularization (AUR) module. The symbols \textcircled{c} and \textcircled{$\uparrow$} denote concatenation and upsampling operations, respectively.
}
\label{fig:allnet}
\vspace{-2mm}
\end{figure*}

Recently, the problem of distortion rectification has been renewed by deep learning techniques due to their strong capacity of visual feature representation~\cite{rong2016radial,yin2018fisheyerecnet}. Instead of explicitly detecting the geometric features in fisheye images, these methods rectify images from learned semantic cues.
Although significant improvements have been obtained, the deep models trained only with semantic-level supervision are prone to overfitting in real-world scenarios with severe distortions due to the remaining gap between the semantics and the geometric nature of distortion rectification.

Regardless of the difficulties of detecting geometric structures in fisheye images, the explicit scene geometries~\cite{devernay2001straight, melo2013unsupervised, zhang2015line} are still strong constraints to rectify distorted images.
Specifically, as a kind of fundamental geometric primitive, straight line segments are very common at the object boundaries of natural images and are also widely used in self-calibration.
As shown in Fig.~\ref{fig:map}, the (distorted) straight lines in (fisheye) images provide reliable cues for fisheye image rectification.
Thus, it is of great interest to investigate how to apply the fundamental geometric properties of straight lines to rectify fisheye images.
In this paper, we follow the classical observation~\cite{devernay2001straight} for camera calibration and distortion rectification that \emph{straight lines in 3D space should remain straight in image planes} to study the problem of fisheye image rectification.
More precisely, motivated by the recent advances of deep learning in edge detection~\cite{XieT17,ManinisPAG18,LiuCHBZBT19}, line segment detection~\cite{huang2018learning,xue-lsd-pami,Xue-HAWP20}, and fisheye image rectification~\cite{rong2016radial,yin2018fisheyerecnet}, we aim to design a
deep-learning-based solution to tackle the problem of fisheye image rectification by explicitly exploiting the line geometry presented in fisheye images.

\subsection{Method Overview}
We approach the problem in the following aspects:
\begin{itemize}
	\item How to design a deep model, \eg,~a convolutional neural network (ConvNet), to fully impose the constraints from the observation of straight lines for fisheye distortion rectification?
	\item How to efficiently train such a deep model for fisheye distortion rectification?
\end{itemize}

For the first problem, we propose a \emph{line-aware rectification network} (LaRecNet) that consists of three major components, as shown in Fig.~\ref{fig:allnet}:
(1) A {\em distorted line perception} (DLP) module takes a fisheye image as input and outputs a heatmap of distorted lines presented in fisheye images;
(2) A {\em multiscale calibration module} is proposed to estimate the distortion parameters from the learned heatmaps of distorted lines and the image appearance;
(3) With the learned distortion parameters, {\em a differentiable rectification layer} is used to rectify the image as well as the learned heatmaps of the distorted lines.
Considering the multiple-task nature of the proposed LaRecNet, we also propose an attentive uncertainty regularization (AUR) module to compactly couple the three major components in the training phase. In detail, the AUR module dynamically estimates the uncertainty maps of the input fisheye images to guide the network to robustly learn the relationship between rectified images and the input images.
At the testing step, given a fisheye image, the distortion parameters are estimated, and the image is rectified in single forward-pass computation.

To train the proposed LaRecNet, we need to feed supervision signals to each component of the network. Specifically, LaRecNet requires the groundtruth of the following as supervision: (1) fisheye images and their expected rectifications, (2) the distortion parameters, and (3) the heatmaps of the distorted lines and rectified lines.
However, to our knowledge, there is no such dataset simultaneously containing these different kinds of data samples. In addition, it is extremely expensive to obtain a large number of precisely annotated distortion parameters for different fisheye cameras.

To this end, we create a {\em synthetic line-rich fisheye} (SLF) dataset that contains all the abovementioned kinds of data annotations for fisheye distortion rectification.
More precisely, our proposed dataset is sourced from two existing datasets, \ie, the wireframe dataset~\cite{huang2018learning} and the SUNCG~\cite{song2016ssc}.
For the base wireframe dataset~\cite{huang2018learning} that has $5462$ images in man-made scenes with line segment annotations,
we randomly generate fisheye projection models with different parameters and apply them to both the images and their annotations to simulate fisheye images and the corresponding distorted lines.
To better depict the imaging process of fisheye cameras, we further enlarge the dataset by rendering the 3D models of the SUNCG dataset~\cite{song2016ssc} with real fisheye lenses in 3D virtual spaces.
With the help of the line-rich characteristics of the proposed SLF dataset, we are able to train the proposed LaRecNet and finally approach the problem of fisheye distortion rectification by explicitly exploiting the line geometry. Moreover, the proposed SLF dataset provides a geometric-toward
way to evaluate the rectification methods.

In the experiments, we demonstrate the effectiveness of our proposed method for fisheye distortion rectification.
Qualitatively, LaRecNet rectifies the distorted lines to be straight even in images with severe geometric distortions.
Quantitatively, our proposed method achieves state-of-the-art performance on several evaluation metrics. In particular, the average reprojection error (RPE) between the corrected images obtained by LaRecNet and the groundtruth images is $0.33$ pixels on the SLF dataset, and the images rectified by LaRecNet achieve the highest image quality of all tested approaches
in both the peak signal-to-noise ratio (PSNR) and structure similarity index (SSIM).
Compared with the preliminary version of our work~\cite{xue2019learning}, we exploit an AUR module to compactly couple the different modules of LaRecNet in the training phase. The enhanced version of our proposed method obtains 1.6\% and 3.4\% relative improvements on the PSNR and SSIM metrics, respectively, while reducing the RPE from $0.48$ pixels to $0.33$ pixels.
A systematic ablation study is performed to further justify the proposed method.

\subsection{Our Contributions}
Our work is distinguished by the following three aspects of fisheye distortion rectification:
\begin{itemize}
	\item
	We propose an end-to-end deep model to impose deep line constraints onto the rectification process of geometric distortions in fisheye images, which achieves state-of-the-art performance on several evaluation metrics.
	
	\item
	In the learning process of distortion rectification, we propose a multiscale calibration block to balance the nonlinear distribution of the geometric distortions. Moreover, the proposed attentive uncertainty regularizer exploits the attention mechanism of distortion rectification to obtain an uncertainty map for the initial rectification, which further improves the performance of our method.
	
	\item
	We propose a large-scale fisheye image dataset to train deep networks and evaluate the effectiveness and efficiency of distortion rectification for fisheye images with line geometry.
	
\end{itemize}

The remainder of this paper is organized as follows.
Sec.~\ref{sec:related-work} briefly recalls the research related to our work. Sec.~\ref{sec:preliminary} introduces the preliminary knowledge of fisheye projection models. Based on the general projection model of the fisheye lens, we present the technical details of our proposed LaRecNet in Sec.~\ref{sec:method}, and the SLF dataset is presented in Sec.~\ref{sec:SLF-dataset}. The experimental results and comparisons are given in Sec.~\ref{sec:experiments}. Finally, we draw some conclusions in Sec.~\ref{sec:conclusion}.

\section{Related Work}
\label{sec:related-work}
\subsection{Distortion Rectification in Digital Images}
The classic pipeline for rectifying geometric distortions in images often involves the following steps: (1) seeking a suitable camera model to characterize the geometric distortions; (2) estimating the parameters to fully depict image formation; and (3) rectifying the distorted images according to the estimated camera model. 
In this section, we recall the related works on distortion rectification in the abovementioned aspects.

\paragraph*{Camera Models with Distortions} 

In early work, \eg,~\cite{snyder1997flattening}, geometric distortions were modeled by derivations of the pinhole camera model. These models can effectively deal with geometric distortions from cameras with small FoVs, while they always fail to handle the cases of fisheye cameras that have large FoVs. Considering that fisheye lenses are designed to capture the entire hemisphere in front of the camera by using special mapping functions~\cite{miyamoto1964fish}, generic camera models have been proposed to approximate such mapping functions. Specifically, Kannala and Brandt~\cite{kannala2006generic} presented a polynomial camera model to approximate the projection of real lenses. Subsequently, Tardif~\emph{et al.}~\cite{TardifSTR09} used the distortion center and distortion circle to model cameras with radially symmetric distortion. Although the distortions of real lenses can be characterized better with more advanced camera models, the large number of parameters and the nonlinearity involved in the camera models often lead to difficulties in the parameter estimation. 
In this paper, we design a deep rectification layer to estimate the parameters of the polynomial generic camera model~\cite{kannala2006generic}. As we shall see, the proposed rectification layer is differentiable and enables us to estimate the distortion parameters and rectify fisheye images in an end-to-end manner. 

\paragraph*{Parameter Estimation for Calibration}
Given a specified camera model, one needs to estimate its parameters to fully characterize the imaging formation. The existing methods for parameter estimation can be approximately  classified into two categories: (1) manual calibration methods with known 3D scene information and (2) self-calibration methods relying on the geometric properties of images. Manual calibration methods can estimate the parameters accurately, \eg,~\cite{heikkila2000geometric,zhang2000flexible,weng1992camera, kannala2006generic,TardifSTR09}; however, they require an expensive calibration apparatus and elaborate setup to ensure that the 3D information of scenes is sufficiently accurate.
To make the calibration process more flexible, some self-calibration methods~\cite{stein1997lens, Faugeras1992Camera,Maybank1992A}
have aimed to use point correspondences and multiview geometry to estimate the internal parameters and external camera motions without knowing the 3D information of scenes. Different from those methods with multiview images, the self-calibration approaches with single-view images attempted to utilize geometric features (\eg, straight lines, conics) of images to recover the internal parameters~\cite{devernay2001straight,zhang2015line,barreto2005geometric,melo2013unsupervised,bukhari2013automatic}. Specifically, a pioneering work~\cite{devernay2001straight} proposed that
the straight line segments in the real world should maintain
their line property even after the projection of the fisheye lens.
Along this axis, Bukhari~\etal~\cite{bukhari2013automatic} recently proposed using an extended Hough transform of lines to correct radial distortions. With a similar assumption,  ``plumb-lines" have also been used to rectify geometric distortions in fisheye images~\cite{zhang2015line, melo2013unsupervised, aleman2014automatic}.
Although these types of calibration methods are simple and effective, their performances depend heavily on the accuracy of geometric object detection results. Our work in this paper
follows the same observations as suggested in~\cite{devernay2001straight}, while we
propose a deep ConvNet to handle the
aforementioned problems and and perform more accurate distorted line extraction in fisheye images.

\subsection{Distortion Rectification with Deep Learning}
Recently, the problem of rectifying geometric distortions in fisheye images with the single-view setting has been advanced by deep-learning-based methods~\cite{rong2016radial,yin2018fisheyerecnet}. Rong~\etal~\cite{rong2016radial} employed ConvNets to regress the distortion parameters from the input distorted images and then used the estimated parameters to rectify images. Subsequently, FishEyeRecNet~\cite{yin2018fisheyerecnet} introduced scene parsing semantics into the rectification network and enabled ones to rectify images in an end-to-end manner. Although some promising results have been reported by these approaches, it is not clear which kind
of high-level geometric information learned from their networks are important for fisheye image rectification. 
In this paper, we aim to open the black box of previous deep-learning-based rectification methods by incorporating straight lines as an explicit constraint that is common to see in manmade environments. Compared with the previous methods that use only deep features without learning geometry, our proposed method significantly improves the performance of fisheye image rectification.

\section{Preliminary to Distortion Rectification}\label{sec:preliminary}
\begin{figure}[tb!]
	\centering
	\includegraphics[width=0.72\linewidth]{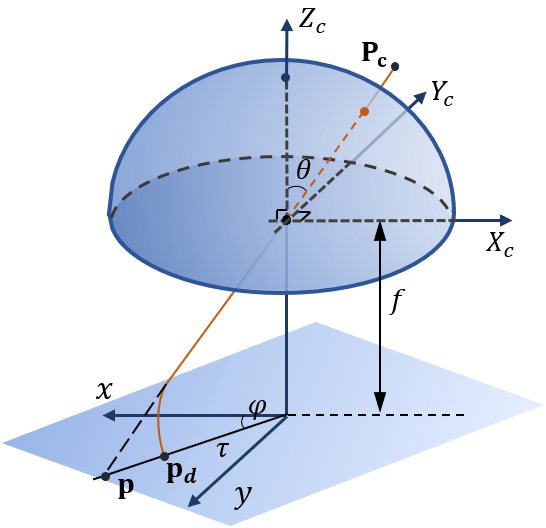}
 	\vspace{-1mm}
	\caption{Camera projection. $\mathbf p$ and $\mathbf{p}_d$ are the projection points of the point $\mathbf P_c$ through pinhole and fisheye lens respectively.}
 	\vspace{-0mm}
	\label{fig:fishproject}
\end{figure}

In this paper, we use a radially symmetric generic camera model~\cite{kannala2006generic} to depict the distortions of fisheye images.
As shown in Fig.~\ref{fig:fishproject}, suppose we have a fisheye camera and a pinhole camera located at the original point in the 3D world with the same orientation, principle point $(u_0,v_0)^T$ and number of pixels per unit distance $m_u, m_v$ in the $x$ and $y$ directions, respectively. The fisheye camera follows the generic camera model~\cite{kannala2006generic}, 
\begin{align}
\label{eq:generic-camera-model}
    \mathcal{R}_f(\theta) = \sum_{i=1}^n k_i\theta^{2i-1},~ n = 1,2,3,4,\ldots,
\end{align}
where $\{k_i\}_i$ are the parameters for the camera model, and $\theta$ is the angle between the incoming ray and the principle axis of the camera. As reported in~\cite{kannala2006generic}, the image formation of a fisheye lens can be approxmated by this model when $n$ reaches $5$. Therefore, we take the sum of the first five polynomial terms as our final fisheye projection model in this paper.

A pinhole camera with focal length $f$ follows the perspective projection model
\begin{align}
\mathcal{R}_p(\theta) = f\tan\theta.
\end{align}
With different projection models, a 3D scene point $\mathbf{P}_c = (X,Y,Z)$ passes through the optical center and yields two image points $\mathbf{p}_d = (x_d,y_d)^T\in\mathbb{R}^2$ by the fisheye camera and $\mathbf{p}=(x,y)^T\in\mathbb{R}^2$ by the pinhole camera, denoted by
\begin{align}
\label{eq:projection-equation}
\begin{split}
    \mathbf{p} &= \mathcal{R}_p(\theta)(\cos\phi,\sin\phi)^T,\\
    \mathbf{p}_d &= \mathcal{R}_f(\theta)(\cos\phi,\sin\phi)^T,
\end{split}
\end{align}
where $(\theta, \phi)^T$ is the direction of the incoming ray, with $\phi$ as the angle between the $x$-axis and the incoming ray. 
Assuming that the pixel coordinate system is orthogonal, we obtain the pixel coordinate $\mathbf{p}_f=(u,v)^T$ in the fisheye image converted from the image coordinate $\mathbf{p}_d$ as 
\begin{align}
    \begin{pmatrix}
    u\\v
    \end{pmatrix}
     = \begin{pmatrix}
    m_u & 0\\0 & m_v
    \end{pmatrix} \mathbf{p}_d + 
    \begin{pmatrix}
    u_0\\v_0
    \end{pmatrix},
\end{align}
where $m_u, m_v$ are the number of pixels per unit distance in the $x$ and $y$ directions, respectively.

Eqn.\,\eqref{eq:projection-equation} implies that a point $\mathbf{p}=(x,y)^T$ in the image plane of the pinhole camera model conveys the incoming ray with direction $(\theta,\phi)^T$ from the 3D scene point to the optical center of the camera. With the direction  $(\theta,\phi)^T$ of the incoming ray, it is straightforward to obtain the corresponding location in the fisheye image plane. 

Thus, with the projection models of fisheye cameras and the target perspective camera available, the geometric distortions in fisheye images can be rectified according to Eqn.~\eqref{eq:projection-equation} once the distortion parameters 
$$\mathbf{k} = (k_1,k_2,k_3,k_4,k_5,m_u,m_v,u_0,v_0)^T$$
are estimated.
For any given fisheye image, we follow this fisheye projection model and learn the parameters $\mathbf{k}$ for rectification.  Notably, this setting is simple and considers mainly the radially symmetric distortion, while as shown in the experimental section, it can suitably handle most of the geometric distortions in fisheye images.

\section{Line-aware Rectification Network}
\label{sec:method}
This section presents the proposed LaRecNet, which aims to approximate the process of rectifying geometric distortions in fisheye images by learning to calibrate straight lines in fisheye images via ConvNets.
\begin{figure*}[t!]
    \centering
    \includegraphics[width=0.87\linewidth]{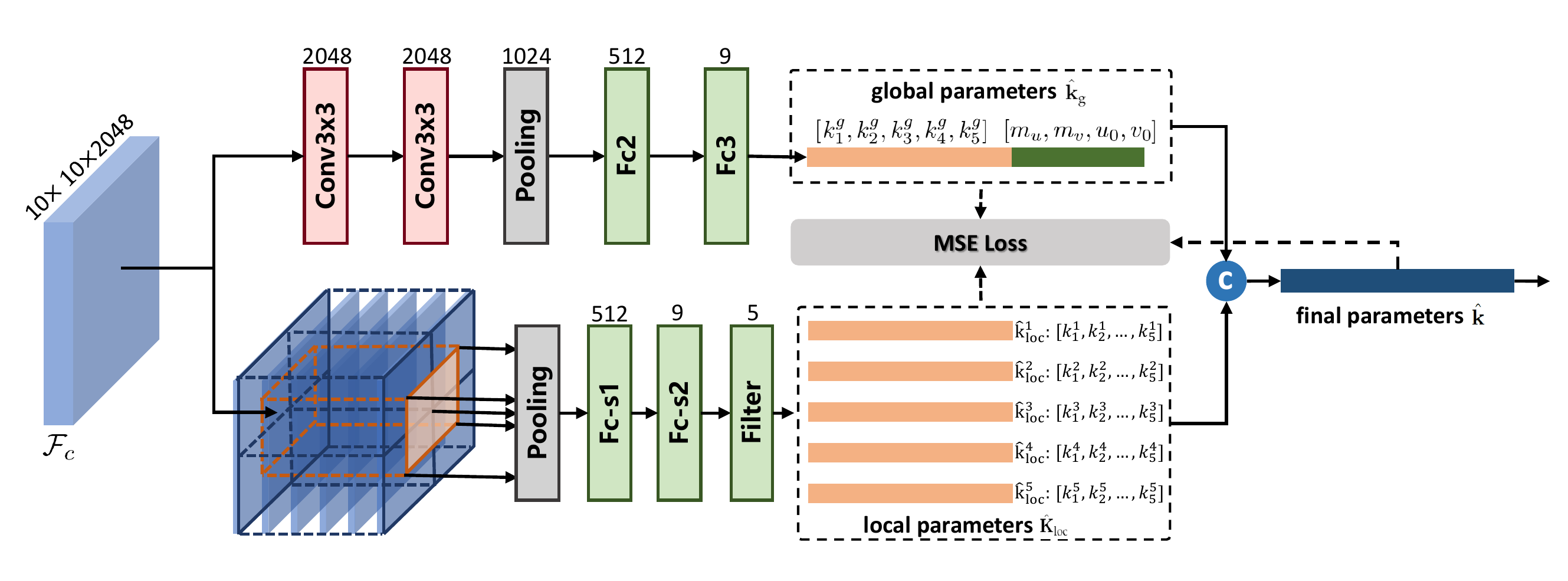}
    \vspace{-3mm}
    \caption{The proposed multiscale calibration module. The final distortion parameters can be regressed through the global and local paths, which use the output feature map $\mathcal{F}_c$ of the previous \emph{Resnet} module as the input.}
    % \caption{Illustration of the proposed multi-scale calibration module. Take the output feature map $\mathcal{F}_s$ of the previous Resnet layer as the input, the global and local distortion parameters are regressed respectively. }
    \label{fig:genenet}
    \vspace{-2mm}
\end{figure*}

\subsection{Network Architecture}
As illustrated in Fig.~\ref{fig:allnet}, the proposed LaRecNet is composed mainly of a {\em distorted line perception module}, a {\em multiscale calibration module}, and a {\em rectification layer}.
Thus, given an input fisheye image $I_f: \Omega \mapsto \mathbb{R}^{3}$ in RGB-color with size $H \times W$ on the image lattice $\Omega = \{0, 1, \cdots, H-1\} \times \{0, 1, \cdots, W-1\}$, LaRecNet yields three outputs: (1) the heatmaps of distorted lines, $\hat{L}_f \in \mathbb{R}^{H\times W}$; (2) the estimated distortion parameters, $\hat{\mathbf{k}}$; and (3) the rectified RGB image, $I_r \in R^{H \times W \time 3}$.

\paragraph*{Distorted Line Perception (DLP) Module}
Inspired by~\cite{devernay2001straight}, we propose learning the rectification process by exploiting straight lines that are highly relevant to geometric distortions.
Accordingly, the DLP module is designed to explicitly learn the heatmaps of distorted lines from the hierarchical deep visual representation. 

To learn $\hat{L}_f$, we combine the pyramid residual module (PRM)~\cite{Han_2017_CVPR} and the stacked hourglass module (SHM)~\cite{Newell2016Stacked}.
In detail, as shown in Fig.~\ref{fig:allnet}~(a), we use cascaded PRMs to extract feature maps of size $\frac{H}{4}\times\frac{W}{4}\times 256$ from the input $I_f$, and then connect five SHMs to compute hierarchical visual features for detecting distorted lines. 
Two deconvolution layers and one convolution layer are finally adopted to make the output heatmaps have the same size as the input. Note that the batch normalization and ReLU layers are used for each (de)convolution layer except for the last prediction layer.

Mathematically, the targeted heatmap $\hat L_f$ of the input image $I_f$ is pixelwise defined as
\begin{equation}\label{eq:line-map}
    \hat L_f(\mathbf{p}) = \left\{
    \begin{array}{cl}
    |\, \ell \,| & \text{if}~{d(\mathbf{p}, \ell)} \leq 1, \, \ell = \arg \min_{\ell_i \in \mathbf{L}} d(\mathbf{p}, \ell_i),\\
    0         & \text{otherwise},
    \end{array}
    \right.
\end{equation}
where $\mathbf{L} = \{\ell_1, \dots, \ell_N\}$ denotes the set of {\em distorted lines}, $d(\mathbf{p}, \ell)$ is the Euclidean distance between pixel $\bf p$ and line segment $\ell$, and $|\, \ell \,|$ measures the length of line $\ell$ on which pixel $\mathbf{p}$ is closely located within $1$ pixel. $\hat L_{f}$ can not only indicate if pixel $\mathbf{p}$ passes through a line but also implicitly imply the relationship with the distortion parameters; that is, the more the distorted lines are curved, the more obvious is the distortion effect.

Subsequently, we use the learned $\hat{L}_f$ as a geometric constraint for estimating the distortion parameters. More precisely, we use the predicted $\hat{L}_f$ as an additional channel of the input image $I_f$, and the concatenated tensors are fed into the multiscale calibration block for estimating the distortion parameters.  
Note that unlike other deep-learning-based methods that learn to rectify fisheye images directly from deep features~\cite{rong2016radial,yin2018fisheyerecnet}, we explicitly exploit the constraints from distorted straight lines in fisheye images.
 
\paragraph*{Multiscale Calibration Module}
As shown in Fig.~\ref{fig:allnet}, the multiscale calibration block is initialized by a convolutional calibration backbone that consists of the first four stages (denoted by L1-L4) of ResNet-50~\cite{he2016deep}. The calibration backbone is designed to extract informative visual features from the fisheye images under the learned deep line constraints. The output feature map is denoted by $\mathcal{F}_c\in\mathbb{R}^{h\times w \times c}$, where $h, w$ and $c$ indicate the height, width and number of channels, respectively. 

Considering the nonlinearity of the distortions, 
we employ a multiscale strategy to predict the distortion parameters for different paths. As illustrated in Fig.~\ref{fig:genenet}, two branches are bifurcated from the extracted feature map $\mathcal{F}_c$.
The global branch predicts the distortion parameters from a global perspective with a series of convolution layers and two fully connected (FC) layers, whose output is a $9$-dimensional vector denoted by 
$$\hat{\mathbf{k}}_{\textrm{g}} = (\hat k_1^g, \hat k_2^g, \hat k_3^g, \hat k_4^g, \hat k_5^g, \hat m_u, \hat m_v, \hat u_0, \hat v_0)^T.$$
For the local branch, the feature map $\mathcal{F}_c$ is partitioned into $5$ subfeatures:
the central region with a size of $6\times6\times1024$ and four surrounded regions, \ie,~the upper left, the lower left, the upper right and the lower right, respectively, with the size of $5\times5\times1024$.

Then, these five sets of subfeatures are fed into two FC layers and a linear filter separately to predict the local parameters, denoted by 
$$\hat{\mathbf{K}}_{\textrm{loc}} = \left\{ \hat{\mathbf{k}}_{\textrm{loc}}^i = \left(\hat k_1^i, \hat k_2^i,\ldots, \hat k_5^i \right)^T\right\}_{i=1}^5.$$
Since $m_u, m_v$ and $u_0, v_0$ are related to the entire image, the local branches do not predict these parameters.
The parameter settings of these two FC layers and the pooling layer are the same as those in the global branch, and these five local streams share the same weights. 
The final output distortion parameter $\hat{\mathbf{k}}$ is the fusion of $\hat{\mathbf{k}}_{\textrm{g}}$ and $\hat{\mathbf{k}}_{\textrm{loc}}^i$ as,
$$\hat{\mathbf{k}} = \left( \bar k_1, \ldots, \bar k_5, \hat m_u, \hat m_v, \hat u_0, \hat v_0 \right)^T,$$
where $\bar k_t$ is the average of $\hat k_t^g$ and $\{\hat k_t^i\}_{i=1}^5$ for $t=1,\cdots, 5$.

\paragraph*{Rectification Layer}
This layer serves as a bridge between the distortion parameters and images. In the forward computation, the rectification layer transforms a fisheye image $I_f$ according to the distortion parameter $\mathbf{\hat{k}}$ to yield a rectified image $I_r$.
More precisely, a pixel $\mathbf{p}_f=(u, v)^T$ in $I_f$ and its corresponding pixel $\mathbf{p}_r=(u_r, v_r)^T$ in $I_r$ are related by
\begin{align}\label{eq:rectify}
    \mathbf{p}_f = \mathcal{T}(\mathbf{p}_r;\, {\hat{\mathbf{k}}})=
    \mathcal{R}_f(\theta) \cdot\frac{\mathbf{p}_r}{\left\|\mathbf{p}_r\right\|_2} +(u_0, \, v_0)^T,
\end{align}
where $\mathcal{T}(\,\cdot\,; \, {\mathbf{\hat{k}}} )$ denotes the forward fisheye projection function with parameters ${\mathbf{\hat{k}}}$.

The rectified image $I_r$ can be obtained by using bilinear interpolation according to Eqn.~\eqref{eq:rectify}. 
Due to the differentiability of the bilinear interpolation, it is possible to optimize the distortion parameter 
by computing the loss between the rectified images $I_r$ and the corresponding groundtruth $I_{\textrm{G}}$. 
In the implementation, we calculate the gradient from the loss function with respect to the distortion parameter $\mathbf{k}_d$ based on the chain rule.

\begin{figure*}[thp!]
    \centering
    \includegraphics[width=0.85\linewidth]{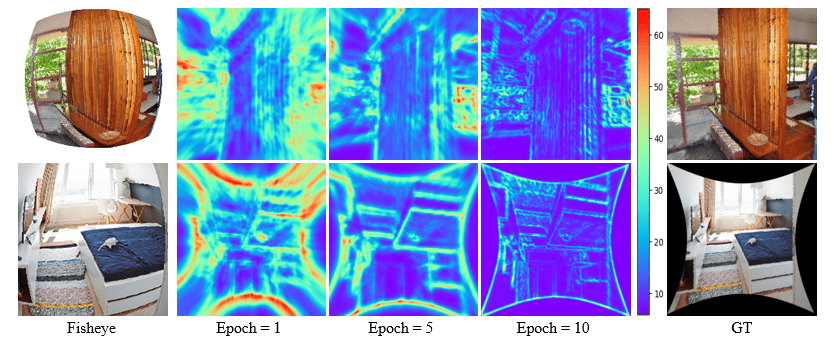}
    \vspace{-3mm}
    \caption{The uncertainty map $U_A$ learned from the attentive regularization module. {\textbf{Left:} RGB fisheye images.} \textbf{Middle:} the learned uncertainty map $U_A$ in different epochs. 
    \textbf{Right:} The groundtruth images.}
    \label{fig:attention}
    \vspace{-1mm}
\end{figure*}

\subsection{Loss Functions and Uncertainty Regularizer}
To train LaRecNet, we employ different loss functions for the several output items. Furthermore, we present an attentive uncertainty regularizer to compactly couple the different components of LaRecNet, which helps the network to focus on the challenging regions for rectification during the training.

\paragraph*{Loss for Learning Distorted Line Heatmaps}
Since the line segments are zero-measure in the image domain, most values of the learning target $\hat{L}_f$ defined in Eqn.~\eqref{eq:line-map} are $0$. 
To avoid the potential sample-imbalance issue between zero-values and nonzero-values of $\hat L_f$, we use a weighted loss function to train the DLP module. 
Specifically, the pixels passing any distorted line segment are collected into the positive set $\Omega^+$, and the remains are collected into the negative set $\Omega^-$, with $\Omega = \Omega^+ \cup \Omega^-$.
Thus, a weighted $l_2$ regression loss can be defined as
\begin{align}
\mathcal{L}_{\textrm{line}} = \frac{|\Omega^-|}{|\Omega|}\sum_{\mathbf{p}\in \Omega^+} D(\mathbf{p}) +  \frac{|\Omega^+|}{|\Omega|} \sum_{\mathbf{p}\in \Omega^-} D(\mathbf{p}), 
\label{eq:line_loss}
\end{align}
where $D(\mathbf{p}) = \|L_{f}(\mathbf{p}) - \hat{L}_f(\mathbf{p})\|_2^2$ with $L_{f}(\mathbf{p})$ as the groundtruth of distorted line segments. 

\paragraph*{Loss for Learning the Distortion Parameters}
The multiscale calibration block outputs a $9$-dimension global parameter $\hat{\mathbf{k}}_{\textrm{g}}$ and $5$ local parameters $\hat{\mathbf{K}}_{\textrm{loc}} = \{\hat{\mathbf{k}}_{\textrm{loc}}^{i} \}_{i=1}^5$ with each $\hat{\mathbf{k}}_{\textrm{loc}}^i $ as a $5$-dimension vector, as well as the fused $9$-dimension parameters $\hat{\mathbf{k}}$. 
We use the MSE loss for training and a weight vector $\bm{\omega}$ of $9$ dimensions to rescale the magnitude between different components of the distortion parameters.
More precisely, given the groundtruth parameters $\mathbf{k}_{\textrm{G}}$, for the output $\hat{\mathbf{k}}_{\textrm{g}}$, we define the loss $\mathcal{L}_{{\textrm{glo}}}$ as
\begin{align}
\mathcal{L}_{\textrm{glo}} = \frac{1}{9} \left\| \bm\omega \cdot (\hat{\mathbf{k}}_{\textrm{g}} - \mathbf{k}_{\textrm{G}}) \, \right\|^2_2.
\label{eq:global}
\end{align}
The loss for the local parameters is defined as
\begin{align}
\mathcal{L}_{\textrm{loc}} = \frac{1}{25} \sum_{i=1}^5 \left\| \bm\omega^{[1:5]}  \cdot ( \hat{\mathbf{k}}_{\textrm{loc}}^i - \mathbf{k}_{\textrm{G}}^{[1:5]}) \right\|^2_2, 
\label{eq:local}
\end{align}
where $\hat{\mathbf{k}}_{\textrm{loc}}^i$ is the $i$-th vector of the predicted local parameters and $\mathbf{a}^{[1:5]}$ is an operator that incorporates the ﬁrst $5$ components of a vector $\mathbf{a}$ as a subvector.
{Similar to the loss $\mathcal{L}_{\textrm{glo}}$, the loss function $\mathcal{L}_{\textrm{fus}}$ is computed to learn the fused distortion parameters $\hat{\mathbf{k}}$. }
The overall loss of the estimation of distortion parameters $\mathcal{L}_{\textrm{para}}$ is
\begin{equation}\label{eq:p_loss}
    \mathcal{L}_{\textrm{para}} = \lambda_{\textrm{fus}}\mathcal{L}_{\textrm{fus}} + \lambda_{\textrm{glo}}\mathcal{L}_{\textrm{glo}} + \lambda_{\textrm{loc}} \mathcal{L}_{\textrm{loc}},
\end{equation}
where $\lambda_{\textrm{fus}}$, $\lambda_{\textrm{glo}}$ and $\lambda_{\textrm{loc}}$ are the weights to balance the different loss items. We set $\lambda_{\textrm{fus}}=2$, $\lambda_{\textrm{glo}}=1$ and $\lambda_{\textrm{loc}}=1$ in our experiments.

\paragraph*{Loss of Geometric Constraints}
Although $\mathcal{L}_{\textrm{para}}$ constrains the network to fit the optimal distortion parameters, the use of $\mathcal{L}_{\textrm{para}}$ alone is susceptible to trapping in a local minimum.
Therefore, considering that the geometric structure can provide a strong constraint to boost performance, 
we design a function $\mathcal{S}_{\hat{\mathbf{k}}, \mathbf{k}_{\textrm{G}}}{(\,\cdot \,)}$ to calculate the geometric errors between the rectified image $I_r$ (using the fused distortion parameters $\hat{\mathbf{k}}$) and the groundtruth image $I_{\textrm{G}}$ (using groundtruth parameters $\mathbf{k}_{\textrm{G}}$), 
\begin{equation}\label{eq:struct loss}
    \mathcal{S}_{\hat{\mathbf{k}}, \mathbf{k}_{\textrm{G}}}{(\mathbf{p}_{f})} =  
     \left\| \mathcal{T}^{-1}(\mathbf{p}_f;\, \hat{\mathbf{k}})-\mathcal{T}^{-1}(\mathbf{p}_f;\, \mathbf{k}_{\textrm{G}}) \right\|_1,
\end{equation}
where $\mathbf{p}_{f}$ is the pixel coordinates of fisheye image $I_f$ and $\mathcal{T}^{-1}(\,\cdot\,; \, {\mathbf{k}} )$ is the inverse of the fisheye projection $\mathcal{T}(\,\cdot\,; \, {\mathbf{k}} )$ described in Eqn.~\eqref{eq:rectify}.

{To make the network focus on the distorted lines, we also use the target heatmap of the distorted lines $\hat{L}_f$ to weight the geometric errors during training. The geometric error of the pixels in $\Omega^+$ should have a large weight because they are more geometrically meaningful than the pixels in $\Omega^-$. Therefore, the total loss function of the geometric constraints is written as 
}
\begin{align}
    \mathcal{L}_{\textrm{geo}} = \frac{\lambda_{\textrm{m}}}{|\Omega|}
     \sum_{\bm{p}_f\in \Omega^+}  
        \mathcal{S}_{\hat{\mathbf{k}}, \mathbf{k}_{\textrm{G}}}{(\mathbf{p}_{f})}  + \frac{1}{|\Omega|} \sum_{\bm{p}_f\in \Omega^-} \mathcal{S}_{\hat{\mathbf{k}}, \mathbf{k}_{\textrm{G}}}{(\mathbf{p}_{f})},
\end{align}
where $\lambda_{\textrm{m}}$ is the weight for positive pixels in the fisheye images. In our experiments, $\lambda_{\textrm{m}}$ is set to 2.

\paragraph*{Attentive Uncertainty Regularization}
% \vspace{-10pt}
Considering that the uncertainty of estimation in the distorted line map $\hat{L}_f$ could result in deviations of distortion parameters and residual distortion in $I_r$, the AUR module is proposed to compensate for the rectification error and achieve a more robust optimization result by exploiting the attention mechanism during rectification.
The proposed AUR module is inspired by the uncertainty estimation of multitask learning~\cite{kendall2018multi} and incorporates the uncertainty into the rectification task to improve the learned heatmaps of distorted lines. 

The architecture of this AUR module is shown in Fig.~\ref{fig:allnet}~(c). Technically, we first select intermediate features from each stage's output of the ResNet (containing the L1-L4 four layers) and then send each of them through a convolution layer with a $1\times 1$ kernel size to downscale the number of channels. Subsequently, bilinear upsampling operations with corresponding scale factors are used to obtain four groups of feature maps of a fixed size: $F_{C_i}\in {R^{H\times W\times C}} (i=1,2,3,4)$.
Then the input RGB fisheye image $I_f$, the generated rectified image $I_r$, and all the multichannel features $F_{C_i}$ are concatenated as a new feature map denoted by $F^{'}_{C}$, which is fed into a convolution layer followed by a softmax activation function to predict the uncertainty map $U_A$:
\begin{equation}
    U_A =\text{Softmax}({F_{C}^{'}}W_A + b_A),
\end{equation}
where $\text{Softmax}(\cdot)$ is the channel-wise softmax function used for normalization. Then, we couple the uncertainty map $U_A$ with the image rectification loss $\mathcal{L}_{\textrm{pix}} = \left\|I_r - I_{\textrm{G}}\right\|_1$ to obtain the attentive regularization as
\begin{equation}
    \mathcal{L}_{\textrm{pix}} \leftarrow \mathcal{L}_{\textrm{pix}}/U_A + \log U_A,
\end{equation}
where $I_r$ is the rectified image and $I_{\textrm{G}}$ is the corresponding groundtruth of $I_r$.

As shown in Fig.~{\ref{fig:attention}}, the uncertainty of all pixels in the rectified image is uniform at the beginning of training. With more epochs being invloved, the number of
pixels far from the geometrically salient regions decreases rapidly. After several epochs, the network focuses on the challenging regions for learning.

\subsection{Training Scheme}\label{sec:training}
The whole training procedure consists of two phases. In the first phase, we optimize the DLP block with the loss function $\mathcal{L}_{\textrm{line}}$ defined in Eqn.~\eqref{eq:line_loss}, 
and we subsequently fix the parameters of the DLP block when it converges.
Then, LaRecNet is trained to learn the distortion parameters and the rectification process in the second phase. The total loss function of the second phase is defined as
\begin{equation}
\mathcal{L} = \lambda_{\textrm{para}}\mathcal{L}_{\textrm{para}}+\lambda_{\textrm{geo}}\mathcal{L}_{\textrm{geo}}+\lambda_{\textrm{pix}}\mathcal{L}_{\textrm{pix}},
\label{eq:loss}
\end{equation}
where $\lambda_{\textrm{para}}$, $\lambda_{\textrm{geo}}$, and $\lambda_{\textrm{pix}}$ are the weights to balance the different terms.
The implementation details are described in Section~\ref{sec:implementation}.

\section{Synthetic Linerich Fisheye (SLF) Dataset}
\label{sec:SLF-dataset}
\begin{figure}[t!]
    \centering
    \includegraphics[width=0.99\linewidth]{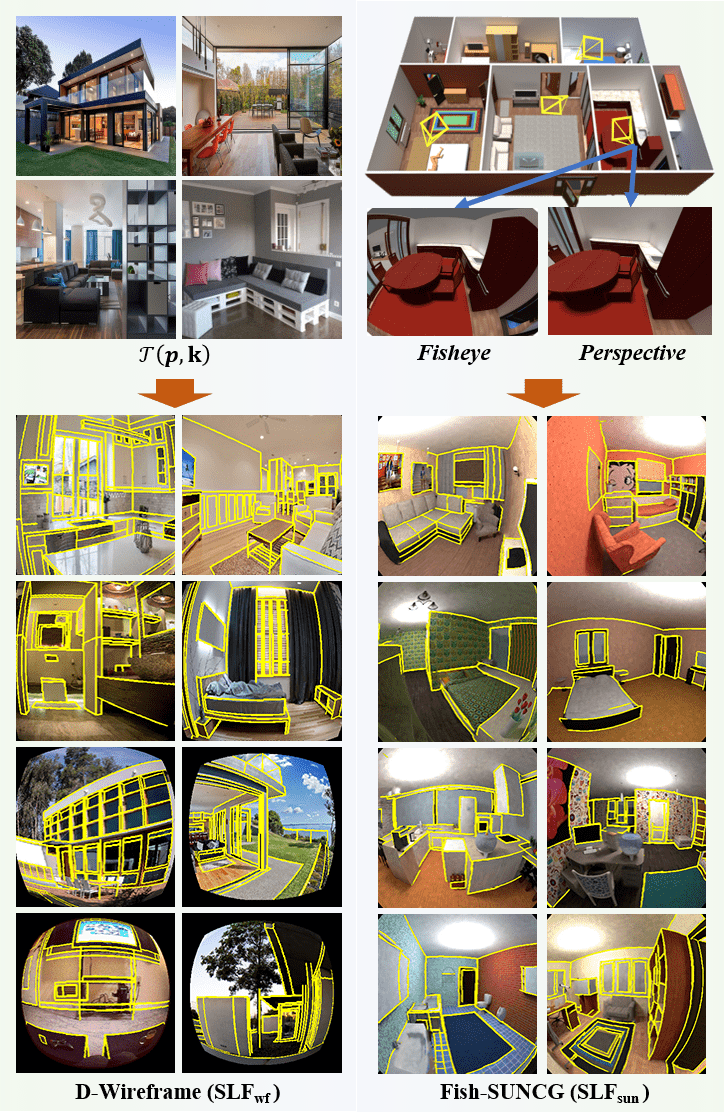}
\vspace{-2mm}
\caption{Data samples from the SLF dataset: D-Wireframe (left) and Fish-SUNCG (right).
	For D-Wireframe, the normal images can be converted to any type of fisheye image using random distortion parameters $\mathbf{k}$; for Fish-SUNCG, given a randomly selected perspective from the virtual rendering scene, the fisheye projection image and the perspective image under this view can be generated simultaneously.}
\label{fig:dataset}
\vspace{-3mm}
\end{figure}

The main difference between our proposed network and previous deep methods for fisheye image rectification is the line constraint, which requires a dataset that contains the annotations of (distorted) line segments for fisheye images and the rectified images, as well as the groundtruth of distortion parameters. However, to our knowledge, there is no such large-scale dataset that can satisfy all the above requirements.
Thanks to the recently released datasets~\cite{huang2018learning,song2016ssc} for geometric vision tasks, we create a new SLF dataset based on the 2D wireframes and 3D surface models of manmade environments for fisheye lens calibration and image rectification. As shown in Fig.~\ref{fig:dataset}, our proposed SLF dataset contains two collections, termed the distorted wireframe collection (D-Wireframe) and the fisheye SUNCG collection (Fish-SUNCG). 
For D-Wiframe, each data point is obtained by distorting the original images and the corresponding line segment annotations with randomly generated distortion parameters. For Fish-SUNCG, we create the data points by simulating the imaging process of fisheye lenses. 

% \vspace{-10pt}

\paragraph*{Distorted Wireframe Collection (D-Wireframe)}
For any image in the wireframes dataset~\cite{huang2018learning}
that contains 5462 normal perspective images and corresponding line segment annotations, we randomly generate $8$ different sets of distortion parameters to obtain fisheye images and the distorted lines according to Eqn.~\eqref{eq:generic-camera-model}. In summary, D-Wireframe contains $41,848$ data samples. According to the original splitting of the wireframe dataset with $5000$ and $462$ images for training and testing, this collection has $40000$ training samples and $1863$ testing samples.

\paragraph*{Fisheye SUNCG Collection (Fish-SUNCG)}
Although D-Wireframe provides many data samples for training, the artificial distortion converted from normal perspective images cannot fully represent the fisheye distortion distributions in the real scenarios. 
To address this problem, we further enrich the dataset by simulating the imaging process of fisheye lenses in the virtual 3D environments provided by SUNCG dataset~\cite{song2016ssc}. The images obtained in this way are collected into Fish-SUNCG.

With the help of the 3D models of SUNCG~\cite{song2016ssc} that contain over 45K virtual 3D scenes with manually created realistic room and furniture layouts, we randomly select a scene in SUNCG and set up a virtual camera equipped with a fisheye and a perspective lens, respectively. 
Then, by using Blender~\cite{blender2014blender}, Fish-SUNCG is built with paired fisheye and perspective images, as shown in Fig.~{\ref{fig:dataset}}.
For line segment generation, we remove the texture of the 3D models to obtain wireframe models of 3D objects, and the 2D wireframe map of the object is generated following perspective imaging.
Then, we manually remove the redundant edges that do not belong to the straight edges of the objects. Finally, paired line maps (the distorted lines in the fisheye image and the straight lines in the perspective image) are generated through the projection transformation between the fisheye lens and the perspective lens.
Since we are able to control the image formation without metric errors, the data samples can be used to train our network without information losses. 
In the end, by rendering in 1,000 scenes, Fish-SUNCG contains 6,000 samples for training and 300 samples for testing. 

In summary, our proposed SLF dataset contains 46,000 training samples and 2,163 testing samples, which consists of two collections: D-Wireframe and Fish-SUNCG, denoted as $\textrm{SLF}_{\textrm{wf}}$ and $\textrm{SLF}_{\textrm{sun}}$, respectively.

\section{Experiments and Analysis}
\label{sec:experiments}
\subsection{Implementation Details}\label{sec:implementation}

Following the training scheme described above, the DLP module is trained first by using the fisheye images and the corresponding heatmaps of distorted lines. Once the DLP module converges, 
we fix the weights of the DLP module and train the remaining components of LaRecNet.
The input size for our network is set as $320\times 320$ for both training and testing.
The weight parameters in Eqn.~\eqref{eq:loss} are set as  $\lambda_{\textrm{geo}} = 100$, $\lambda_{\textrm{pix}}=\lambda_{\textrm{para}}=1$, and the balance parameters are set as $\bm{\omega} =\{ \omega_1=0.1, \omega_2=0.1, \omega_3=0.5, \omega_4=1, \omega_5=1, \omega_6=0.1, \omega_7=0.1, \omega_8=0.1, \omega_9=0.1 \}$. 

For optimization, stochastic gradient descent (SGD) is used to train the network with an initial learning rate $0.001$ in both training phases. The learning rate is decayed by a factor of $10$ after $50$ epochs. 
In the first phase, the DLP module converges after $100$ epochs. In the second phase, the loss function reaches a plateau after $150$ epochs of learning. Our network is implemented with PyTorch~\cite{PyTorch}.

\subsection{Benchmark Datasets}
Our proposed SLF dataset contains two collections: $\text{SLF}_{\text{wf}}$ and $\text{SLF}_{\text{sun}}$. We train the proposed LaRecNet on the training split of the entire SLF dataset as the baseline. Moreover, we train the network on these two collections independently to further justify that these two parts are complementary to achieve better performance. 

For evaluation, we first test our method and the previous state-of-the-art methods on the testing split of the SLF dataset. Then, we test these methods on a public dataset~\cite{fishdataset2016-5} that contains both synthetic and real-world video sequences taken by fisheye cameras. 
We call the dataset proposed in~\cite{fishdataset2016-5} the \emph{fisheye video dataset} for the simplicity of representation.
Finally, we fetch some fisheye images without available groundtruth from the Internet to qualitatively verify the generalization ability of our proposed method.

\subsection{Evaluation Metrics}
Previously, the PSNR and SSIM evaluation metrics were used to quantitatively evaluate the rectified images. These two metrics are usually used for image restoration; however, they cannot precisely measure the geometric accuracy of rectified images. 
To address this problem, we take the distortion parameters into account to design two new metrics to evaluate this accuracy.

\paragraph*{PSNR and SSIM for Rectified Images} 
The two metrics are widely used to describe the degree of pixel blurring and structure distortion, so we use them here to compare the rectified fisheye images with the groundtruth. Larger values of the PSNR and SSIM~\cite{SSIM} indicate better rectification results.

\begin{figure*}[t!]
    \centering
    \subfigure[{Fisheye}]{
    \begin{minipage}[b]{0.14\linewidth}
    \includegraphics[width=1\linewidth]{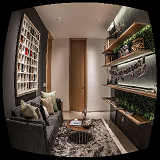}\vspace{1pt}
    \includegraphics[width=1\linewidth]{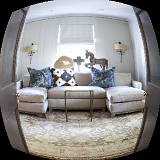}\vspace{1pt}
    \includegraphics[width=1\linewidth]{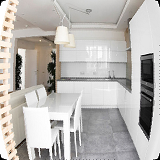}\vspace{1pt}
    \includegraphics[width=1\linewidth]{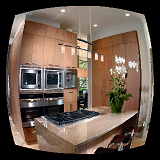}
    \end{minipage}}
    \hspace{-1.6ex}
    \subfigure[{ Distorted Lines}]{
    \begin{minipage}[b]{0.14\linewidth}
    \includegraphics[width=1\linewidth]{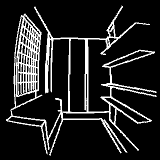}\vspace{1pt}
    \includegraphics[width=1\linewidth]{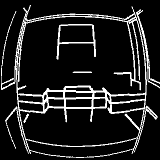}\vspace{1pt}
    \includegraphics[width=1\linewidth]{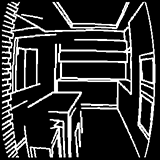}\vspace{1pt}
    \includegraphics[width=1\linewidth]{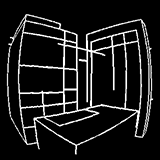}
    \end{minipage}}
    \hspace{-1.6ex}
    \subfigure[{ Bukhari~\cite{bukhari2013automatic}}]{
    \begin{minipage}[b]{0.14\linewidth}
    \includegraphics[width=1\linewidth]{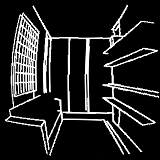}\vspace{1pt}
    \includegraphics[width=1\linewidth]{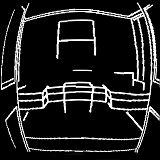}\vspace{1pt}
    \includegraphics[width=1\linewidth]{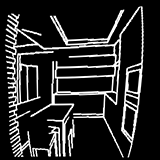}\vspace{1pt}
    \includegraphics[width=1\linewidth]{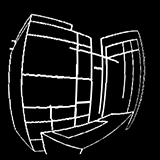}
    \end{minipage}}
    \hspace{-1.6ex}
    \subfigure[{ AlemnFlores~\cite{aleman2014automatic}}]{
    \begin{minipage}[b]{0.14\linewidth}
    \includegraphics[width=1\linewidth]{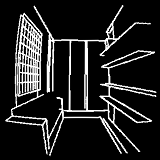}\vspace{1pt}
    \includegraphics[width=1\linewidth]{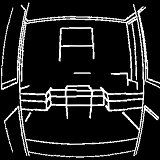}\vspace{1pt}
    \includegraphics[width=1\linewidth]{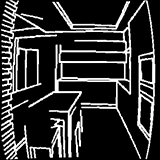}\vspace{1pt}
    \includegraphics[width=1\linewidth]{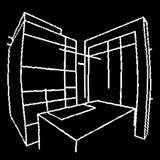}
    \end{minipage}}
    \hspace{-1.6ex}
    \subfigure[{ Rong~\cite{rong2016radial}}]{
    \begin{minipage}[b]{0.14\linewidth}
    \includegraphics[width=1\linewidth]{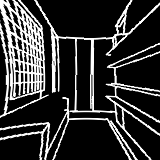}\vspace{1pt}
    \includegraphics[width=1\linewidth]{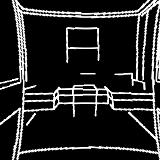}\vspace{1pt}
    \includegraphics[width=1\linewidth]{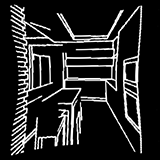}\vspace{1pt}
    \includegraphics[width=1\linewidth]{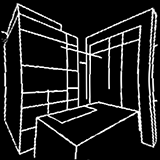}
    \end{minipage}}
    \hspace{-1.6ex}
    \subfigure[{Ours}]{
    \begin{minipage}[b]{0.14\linewidth}
    \includegraphics[width=1\linewidth]{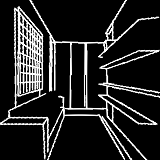}\vspace{1pt}
    \includegraphics[width=1\linewidth]{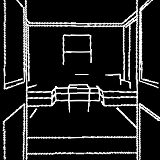}\vspace{1pt}
    \includegraphics[width=1\linewidth]{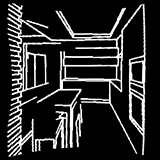}\vspace{1pt}
    \includegraphics[width=1\linewidth]{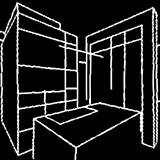}
    \end{minipage}}
    \hspace{-1.6ex}
    \subfigure[{GT}]{
    \begin{minipage}[b]{0.14\linewidth}
    \includegraphics[width=1\linewidth]{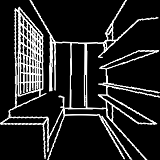}\vspace{1pt}
    \includegraphics[width=1\linewidth]{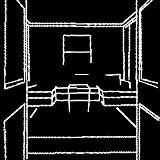}\vspace{1pt}
    \includegraphics[width=1\linewidth]{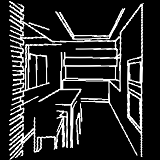}\vspace{1pt}
    \includegraphics[width=1\linewidth]{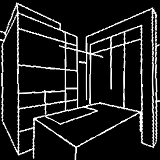}
    \end{minipage}}
    \vspace{-3pt}
    \caption{Distortion line rectification results of various methods. From left to right are the input RGB fisheye images, the distorted lines detected in fisheye images, the rectified results by different methods~{\cite{bukhari2013automatic,aleman2014automatic,rong2016radial}}, our proposed method, and the groundtruths.}
    \label{fig:input_line}
    \vspace{-1mm}
\end{figure*}
\begin{figure}[htp!]
\vspace{-3mm}
\centering
\includegraphics[width=0.9\linewidth]{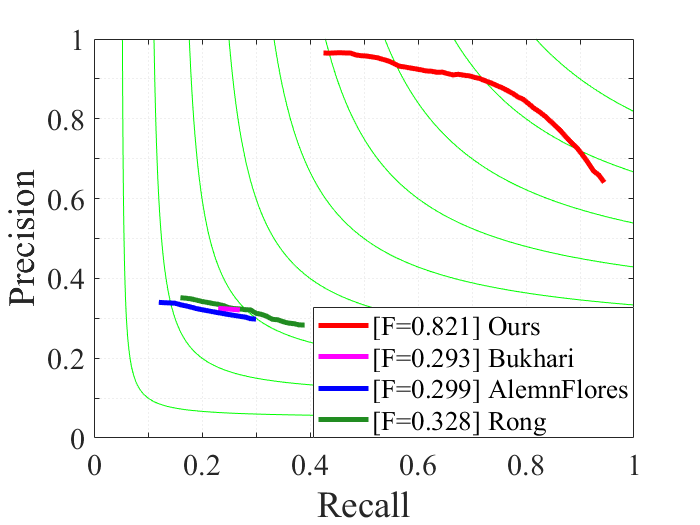}
\vspace{-3mm}
\caption{The precision-recall curves of different rectification methods for the line map rectification~\cite{bukhari2013automatic,aleman2014automatic,rong2016radial}.}
\vspace{-3mm}
\label{fig:PR}
\end{figure}

\paragraph*{Precision-vs-Recall for the Rectified Heatmaps of Lines}
Benefiting from the line-rich characteristics of our proposed SLF dataset, we are able to measure the geometric accuracy by comparing the rectified heatmaps of distorted lines with the groundtruth of line segments. Motivated by the evaluation protocols used for edge detection~\cite{Martin2004Learning} and line segment detection~\cite{huang2018learning,xue-lsd-pami}, we use the precision and recall to measure if the pixels on the distorted lines are still on the straight lines after rectification. Denoting the rectified heatmap of the distorted lines by $\hat{L}_r$ , we use the matching strategy proposed in~\cite{Martin2004Learning} to match the edge pixels with the groundtruth $\hat{L}_{\textrm{G}}$. Then, we calculate the precision and recall by
\begin{equation}
\text{Precision} = |P\cap G|/|P|,~~ \text{Recall} = |P\cap G|/|G|,
\end{equation}
where $P$ is the set of edge pixels in $\hat{L}_r$, $G$ is the set of edge pixels in the groundtruth $\hat{L}_{\textrm{G}}$ and $P\cap G$ is the set of the correctly rectified edge pixels. Since the values of the heatmaps $\hat{L}_r$ and $\hat{L}_{\textrm{G}}$ are lengths of the line segments, we apply a set of thresholds $\tau \in \{5,10,20,40,80,120,160,200,240,255\}$ to the heatmap $\hat{L}_r$ to obtain the edge pixels of the line segments with lengths smaller than $\tau$, which yields a parametric precision-recall curve for evaluation. The overall performance is calculated by the maximal F-score of every pair of precision and recall, denoted by
\begin{equation}
    F = \frac{2\cdot\text{Precision}\cdot\text{Recall}}{\text{Precision}+\text{Recall}}.
\end{equation}

Furthermore, we calculate the RPE of the distance of the rectified pixels of fisheye images to the groundtruth to measure the geometric accuracy of the rectification results.

\paragraph*{Reprojection Error (RPE)}
The RPE is usually used to quantify the distance between an estimation of a 2D/3D point and its true projection position. Given the real distortion parameters $\mathbf{k}$ and the estimated ones $\mathbf{\hat{k}}$, we rectify any pixel $\mathbf{p}_f$ of the fisheye image and calculate the RPE as
\begin{equation}
    \gamma(\mathbf{p}_f; \mathbf{k}, \mathbf{\hat{k}}) = \left\|\mathcal{T}^{-1}(\mathbf{p}_f; \mathbf{k}) - \mathcal{T}^{-1}(\mathbf{p}_f; \mathbf{\hat{k}})\right\|_2,
\end{equation}
where $\mathcal{T}^{-1}$ maps a pixel of a fisheye image into the domain of the rectified image. The averaged RPE over all the pixels is used to measure the geometric accuracy for the estimated distortion parameters.

\subsection{Main Results}
In this section, we compare the proposed LaRecNet with the state-of-the-art methods~\cite{bukhari2013automatic,aleman2014automatic,rong2016radial} on our proposed SLF dataset, the fisheye video dataset~\cite{fishdataset2016-5} and the images fetched from the Internet. More experimental results can be seen in \url{https://xuezhucun.github.io/LaRecNet}.

\subsubsection{Results on the SLF Dataset}
\begin{figure*}[thp!]
    \centering
    \subfigure[{ Fisheye}]{
    \begin{minipage}[b]{0.15\linewidth}
    \includegraphics[width=1\linewidth]{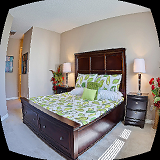}\vspace{1pt}
    \includegraphics[width=1\linewidth]{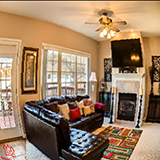}\vspace{1pt}
     \includegraphics[width=1\linewidth]{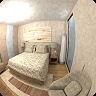}\vspace{1pt}
    \includegraphics[width=1\linewidth]{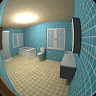}
    \end{minipage}}
    \hspace{-1.6ex}
    \subfigure[{ Bukhari~\cite{bukhari2013automatic}}]{
    \begin{minipage}[b]{0.15\linewidth}
    \includegraphics[width=1\linewidth]{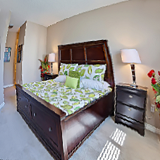}\vspace{1pt}
    \includegraphics[width=1\linewidth]{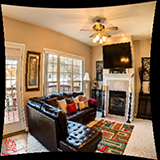}\vspace{1pt}
     \includegraphics[width=1\linewidth]{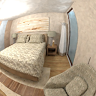}\vspace{1pt}
    \includegraphics[width=1\linewidth]{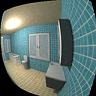}
    \end{minipage}}
    \hspace{-1.6ex}
    \subfigure[{ AlemnFlores~\cite{aleman2014automatic}}]{
    \begin{minipage}[b]{0.15\linewidth}
    \includegraphics[width=1\linewidth]{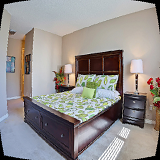}\vspace{1pt}
    \includegraphics[width=1\linewidth]{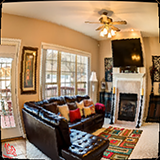}\vspace{1pt}
    \includegraphics[width=1\linewidth]{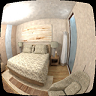}\vspace{1pt}
    \includegraphics[width=1\linewidth]{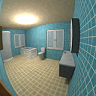}
    \end{minipage}}
    \hspace{-1.6ex}
    \subfigure[{ Rong~\cite{rong2016radial}}]{
    \begin{minipage}[b]{0.15\linewidth}
    \includegraphics[width=1\linewidth]{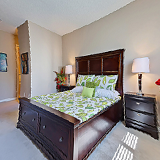}\vspace{1pt}
    \includegraphics[width=1\linewidth]{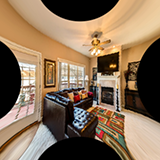}\vspace{1pt}
    \includegraphics[width=1\linewidth]{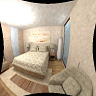}\vspace{1pt}
    \includegraphics[width=1\linewidth]{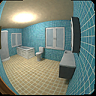}
    \end{minipage}}
    \hspace{-1.6ex}
    \subfigure[{ Ours}]{
    \begin{minipage}[b]{0.15\linewidth}
    \includegraphics[width=1\linewidth]{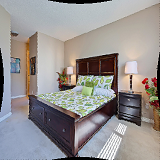}\vspace{1pt}
    \includegraphics[width=1\linewidth]{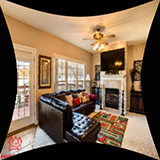}\vspace{1pt}
    \includegraphics[width=1\linewidth]{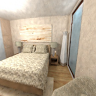}\vspace{1pt}
    \includegraphics[width=1\linewidth]{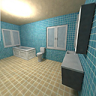}
    \end{minipage}}
     \hspace{-1.6ex}
    \subfigure[{ GT}]{
    \begin{minipage}[b]{0.15\linewidth}
    \includegraphics[width=1\linewidth]{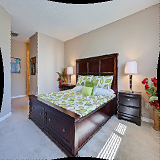}\vspace{1pt}
    \includegraphics[width=1\linewidth]{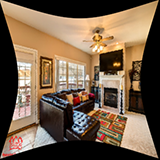}\vspace{1pt}
    \includegraphics[width=1\linewidth]{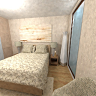}\vspace{1pt}
    \includegraphics[width=1\linewidth]{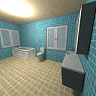}
    \end{minipage}}
    \vspace{-3mm}
    \caption{Qualitative comparison results of fisheye image rectification on $\text{SLF}_{\text{wf}}$ and $\text{SLF}_{\text{sun}}$. From left to right are the input fisheye images, rectification results of three state-of-the-art methods (Bukhari~\cite{bukhari2013automatic}, AlemnFlores~\cite{aleman2014automatic}, Rong~\cite{rong2016radial}), our results and the groundtruth images.}
    \vspace{-5mm}
    \label{fig:syn}
\end{figure*}

We report the quantitative evaluation results of the previous state-of-the-art methods and the proposed LaRecNet in Tab.~\ref{tab:synthetic}. Since the proposed SLF dataset has two collections %$\mathcal{D}_{wf}$ and $\mathcal{D}_{sun}$
$\text{SLF}_{wf}$ and $\text{SLF}_{sun}$, we additionally train LaRecNet on $\text{SLF}_{wf}$ and $\text{SLF}_{sun}$, respectively. 
For FishRectNet~\cite{yin2018fisheyerecnet}, we use the evaluation results of the PSNR and SSIM presented in their paper for comparison, which is associated with the nonpublic source codes and datasets of the project.
According to the results presented in Tab.~\ref{tab:synthetic} and Figs.~\ref{fig:input_line} to~\ref{fig:syn}, 
our proposed method obtains significant improvements in all the evaluation metrics, regardless of the training set we selected. 
In the next  we discuss the experimental results on two aspects of the geometric accuracy and the image quality of the rectification results.

\begin{figure*}[thp!]
    \centering
    \subfigure[{Fisheye}]{
    \begin{minipage}[b]{0.15\linewidth}
    \includegraphics[width=1\linewidth]{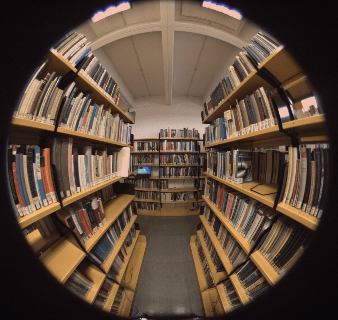}\vspace{1pt}
    \includegraphics[width=1\linewidth]{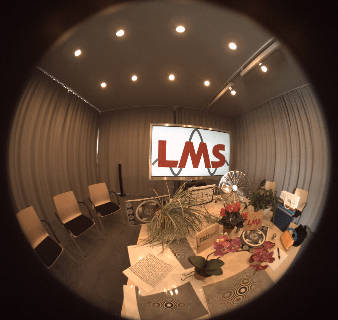}\vspace{1pt}
    \includegraphics[width=1\linewidth]{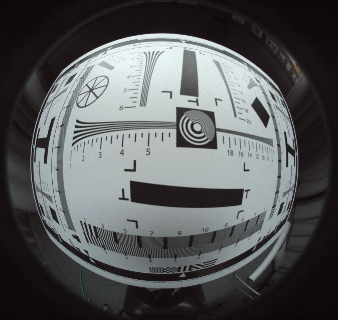}\vspace{1pt}
    \includegraphics[width=1\linewidth]{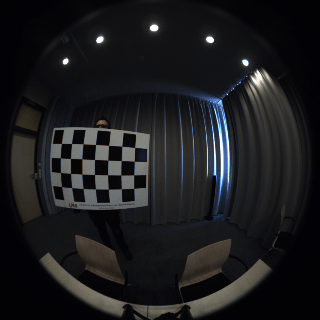}
    \end{minipage}}
    \hspace{-1.6ex}
    \subfigure[{Bukhari~\cite{bukhari2013automatic}}]{
    \begin{minipage}[b]{0.15\linewidth}
    \includegraphics[width=1\linewidth]{exp-real/LibraryA_0001.png}\vspace{1pt}
    \includegraphics[width=1\linewidth]{exp-real/ClutterA_0001.png}\vspace{1pt}
    \includegraphics[width=1\linewidth]{exp-real/TestchartA_0001.png}\vspace{1pt}
    \includegraphics[width=1\linewidth]{exp-real/Calibration13.png}
    \end{minipage}}
    \hspace{-1.6ex}
    \subfigure[{AlemnFlores~\cite{aleman2014automatic}}]{
    \begin{minipage}[b]{0.15\linewidth}
    \includegraphics[width=1\linewidth]{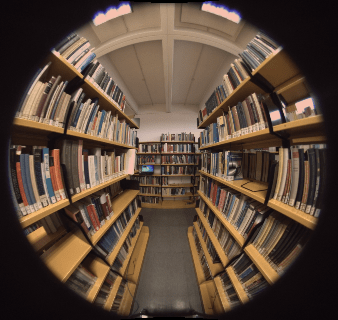}\vspace{1pt}
    \includegraphics[width=1\linewidth]{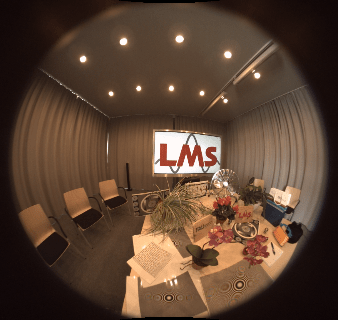}\vspace{1pt}
    \includegraphics[width=1\linewidth]{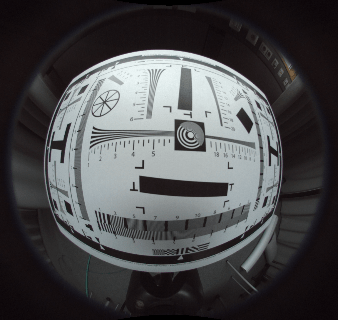}\vspace{1pt}
    \includegraphics[width=1\linewidth]{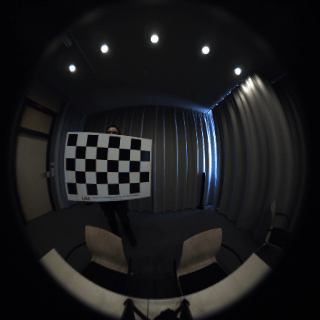}
    \end{minipage}}
    \hspace{-1.6ex}
    \subfigure[{Rong~\cite{rong2016radial}}]{
    \begin{minipage}[b]{0.15\linewidth}
    \includegraphics[width=1\linewidth]{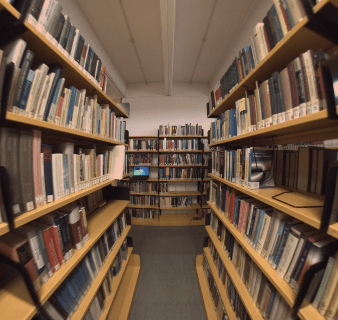}\vspace{1pt}
    \includegraphics[width=1\linewidth]{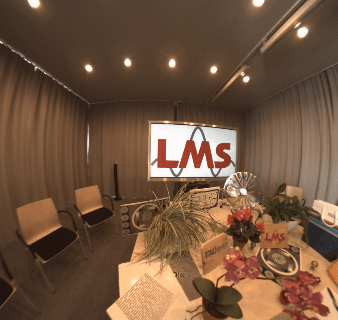}\vspace{1pt}
    \includegraphics[width=1\linewidth]{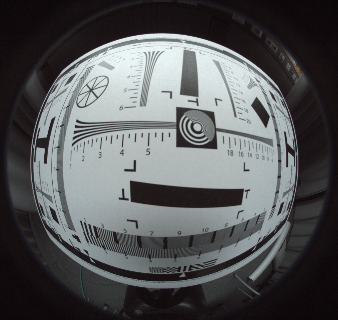}\vspace{1pt}
    \includegraphics[width=1\linewidth]{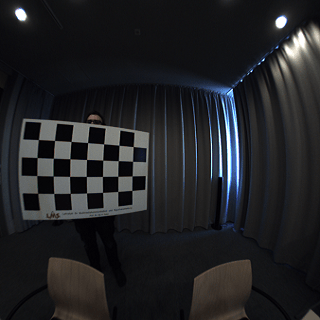}
    \end{minipage}}
    \hspace{-1.6ex}
    \subfigure[{Ours}]{
    \begin{minipage}[b]{0.15\linewidth}
    \includegraphics[width=1\linewidth]{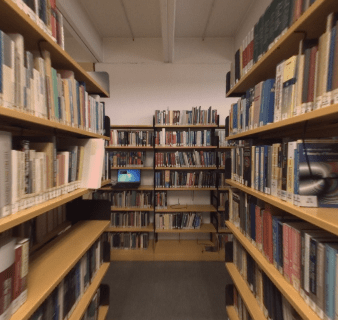}\vspace{1pt}
    \includegraphics[width=1\linewidth]{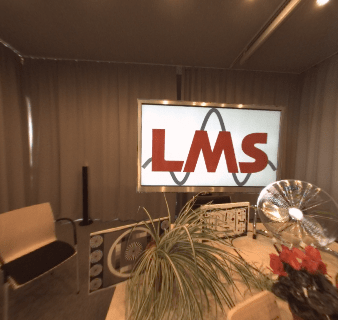}\vspace{1pt}
    \includegraphics[width=1\linewidth]{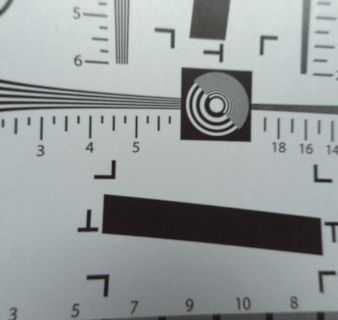}\vspace{1pt}
    \includegraphics[width=1\linewidth]{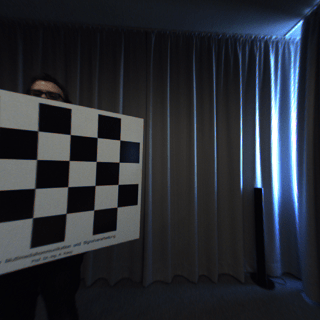}
    \end{minipage}}
    \hspace{-1.6ex}
    \subfigure[{GT}]{
    \begin{minipage}[b]{0.15\linewidth}
    \includegraphics[width=1\linewidth]{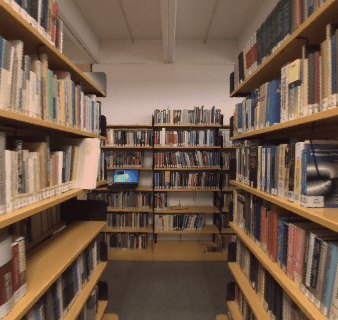}\vspace{1pt}
    \includegraphics[width=1\linewidth]{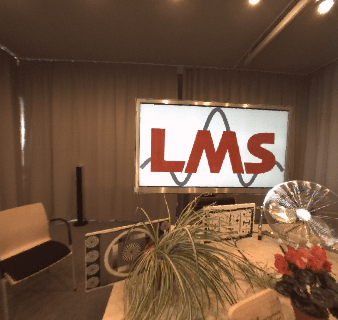}\vspace{1pt}
    \includegraphics[width=1\linewidth]{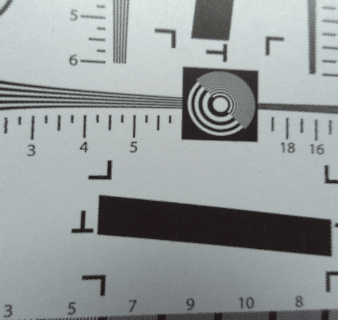}\vspace{1pt}
    \includegraphics[width=1\linewidth]{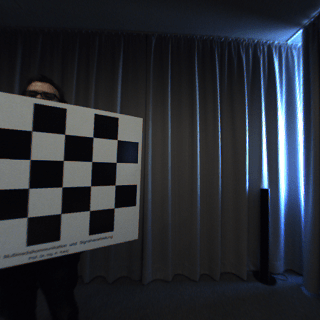}
    \end{minipage}}
    \vspace{-3mm}
    \caption{Qualitative rectification comparison results on a real fisheye dataset~\cite{fishdataset2016-5}. From left to right are the input fisheye images, rectification results of three state-of-the-art methods (Bukhari~\cite{bukhari2013automatic}, AlemnFlores~\cite{aleman2014automatic}, Rong~\cite{rong2016radial}), our results and the groundtruth images.}
    \vspace{-3mm}
    \label{fig:real1}
\end{figure*}

\paragraph*{Geometric Accuracy}
A straightforward way to evaluate the rectified images is to observe whether the distorted lines can be straightened after rectification. Ideally, {\em the projections of straight lines from space to the camera remain straight}.
Accordingly, we use the line segments in the proposed SLF dataset for evaluation. In Fig.~\ref{fig:input_line}, the rectified heatmaps of lines are displayed for qualitative evaluation. Benefiting from the explicit exploitation of the line geometry and end-to-end learning, our method obtains the best performance of all the tested methods
%Editor: Please ensure that the intended meaning has been maintained in this edit.
for rectifying the distorted lines. By contrast, there still exist some obvious distortions in the results of other methods. Moreover, we evaluate the rectification results based on the metrics of the F-score and precision-recall curves of the heatmaps of lines, reported in Fig.~\ref{fig:PR} and Tab.~\ref{tab:synthetic}.
The precision-recall curves show that the rectified distortion lines obtained by our proposed method are the closest to the original geometry relative to other methods in the accuracy as represented by the F-score (F-value=0.821), which also demonstrates that our proposed LaRecNet is far superior to other methods in the aspect of geometric accuracy. Finally, we use the estimated parameters by different methods to calculate the RPEs for every pixel in the fisheye images. As reported in Tab.~\ref{tab:synthetic}, only the proposed method can precisely rectify the pixels in the fisheye images with an error of less than 0.5 pixels.
\begin{table}[h]
	\vspace{-2mm}
	\caption{Quantitative evaluation on the proposed SLF dataset with the metrics of the PSNR, SSIM, F-score and RPE. For our method, the baseline model is trained on all the training samples of the SLF dataset. The models trained on the collections of $\text{SLF}_{\text{wf}}$ and $\text{SLF}_{\text{sun}}$ are indicated with the corresponding suffixes.}
	\vspace{-4mm}
	\label{tab:synthetic}
	\centering
	\begin{tabular}{l|ll|lll}
		\toprule
		Method & PSNR & SSIM & F & RPE  \\ \hline
		Bukhari~\cite{bukhari2013automatic}     & 9.34 & 0.18 & 0.29 & 164.75\\
		AlemnFlores~\cite{aleman2014automatic}  & 10.23 & 0.26 & 0.30 & 125.42\\
		Rong~\etal~\cite{rong2016radial} & 12.92 & 0.32 & 0.33 & 121.69\\
		FishRectNet~\cite{fishdataset2016-5}* & 14.96 & 0.41& N/A & N/A\\ \midrule
		Our method ($\text{SLF}$)   &\textbf{28.06} & \textbf{0.90} &\textbf{0.82}  &\textbf{0.33}\\
		Our method ($\text{SLF}_\text{wf}$) &24.73          & 0.77     &0.81   &0.38  \\
		Our method ($\text{SLF}_\text{sun}$) &15.03       &0.43      &0.30     &25.42  \\ \bottomrule
	\end{tabular}
	\vspace{-1mm}
\end{table}

\paragraph*{Image Quality}
We also evaluate the proposed method by following the previous evaluation protocols used in~\cite{rong2016radial,yin2018fisheyerecnet}, which can reflect the quality of the rectified images.
The results in Tab.~\ref{tab:synthetic} demonstrate that our method is clearly
superior to other methods and achieves the best performance on the PSNR and SSIM metrics. Qualitatively, we display the rectified images on the SLF dataset in Fig.~\ref{fig:syn}. Specifically, we select the testing images with different types of fisheye distortion (\eg, full-frame fisheye images, full circle fisheye images and drum fisheye images) for visualization. The results show that our method accurately rectifies the fisheye images, while other methods cannot handle various types of distortion and obtain incorrect rectification results.

\subsubsection{Results on the Fisheye Video Dataset}
To further justify our proposed method, we use the fisheye video dataset proposed in~\cite{fishdataset2016-5} for evaluation. Note that we only perform the evaluation without tuning the trained models on the SLF dataset.
In this dataset, we perform the comparison only on the metrics of PSNR, SSIM and RPE due to the lack of annotations of distorted lines. To obtain the groundtruth of the fisheye video dataset,
we use the calibration toolbox~\cite{scaramuzza2006toolbox} to estimate internal and external parameters from the video of the calibration pattern in this dataset.
\begin{table}[t!]
	\centering
	\vspace{-2mm}
	\caption{Quantitative evaluation on the \emph{fisheye video dataset}~\cite{fishdataset2016-5} with the metrics of PSNR, SSIM, and RPE. For our method, the baseline model is trained on all the training samples of the SLF dataset. The models trained on the collections of $\text{SLF}_{\text{wf}}$ and $\text{SLF}_{\text{sun}}$ are indicated with the corresponding suffixes. }
	\begin{tabular}{l|llll}
		\toprule
		Method & PSNR & SSIM  & RPE  \\ \hline
		Bukhari~\cite{bukhari2013automatic}     & 9.84 & 0.16 & 156.3\\
		AlemnFlores~\cite{aleman2014automatic}  &10.72 & 0.30 & 125.31\\
		Rong~\etal~\cite{rong2016radial} & 11.81 & 0.30  & 125.31\\ \midrule
		% FishRectNet~\cite{fishdataset2016-5}* & N/A & N/A & N/A\\ \hline
		Our method (\text{SLF})  &\textbf{22.34} & \textbf{0.82} &\textbf{1.68}\\
		Our method ($\text{SLF}_{\text{wf}}$) &19.86         &0.67     &1.91  \\
		Our method ($\text{SLF}_{\text{sun}}$) &11.34       &0.39          &120.12  \\ \bottomrule
	\end{tabular}
	\label{tab:real2}
	\vspace{-2mm}
\end{table}
As shown in Fig.~\ref{fig:real1}, our proposed baseline model can accurately rectify the fisheye video sequences in single forward-pass computation. Quantitatively, we report the PSNR, SSIM and RPE for different methods in Tab.~\ref{tab:real2}. It can be concluded that our proposed method has the best performance in distortion rectification, while other methods cannot robustly rectify the video sequences.

\begin{figure*}[thp!]
	\centering
	\subfigure[{Fisheye}]{
		\begin{minipage}[b]{0.19\linewidth}
			\includegraphics[width=1\linewidth]{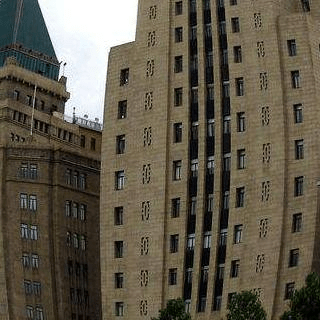}\vspace{2pt}
			\includegraphics[width=1\linewidth]{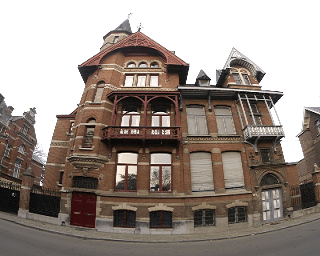}\vspace{2pt}
			\includegraphics[width=1\linewidth]{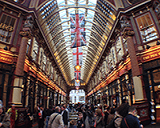}\vspace{2pt}
			\includegraphics[width=1\linewidth]{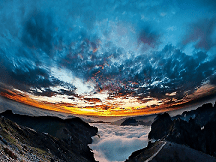}
	\end{minipage}}
	\hspace{-1.5ex}
	\subfigure[{Bukhari~\cite{bukhari2013automatic}}]{
		\begin{minipage}[b]{0.19\linewidth}
			\includegraphics[width=1\linewidth]{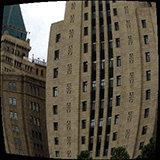}\vspace{2pt}
			\includegraphics[width=1\linewidth]{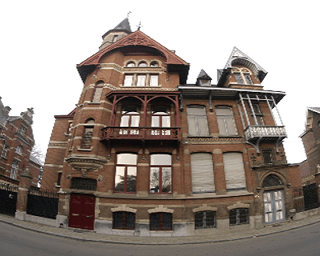}\vspace{2pt}
			\includegraphics[width=1\linewidth]{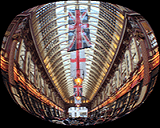}\vspace{2pt}
			\includegraphics[width=1\linewidth]{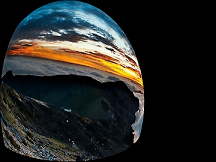}
	\end{minipage}}
	\hspace{-1.5ex}
	\subfigure[{AlemnFlores~\cite{aleman2014automatic}}]{
		\begin{minipage}[b]{0.19\linewidth}
			\includegraphics[width=1\linewidth]{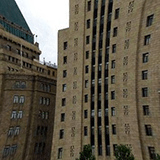}\vspace{2pt}
			\includegraphics[width=1\linewidth]{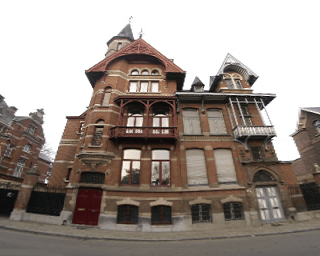}\vspace{2pt}
			\includegraphics[width=1\linewidth]{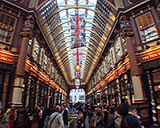}\vspace{2pt}
			\includegraphics[width=1\linewidth]{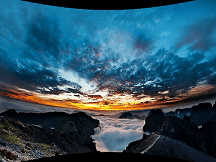}
	\end{minipage}}
	\hspace{-1.5ex}
	\subfigure[{Rong~\cite{rong2016radial}}]{
		\begin{minipage}[b]{0.19\linewidth}
			\includegraphics[width=1\linewidth]{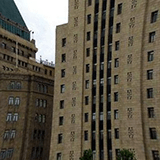}\vspace{2pt}
			\includegraphics[width=1\linewidth]{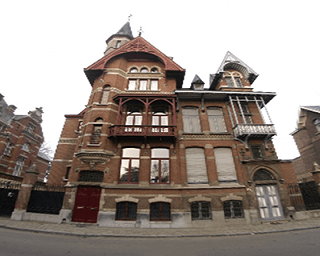}\vspace{2pt}
			\includegraphics[width=1\linewidth]{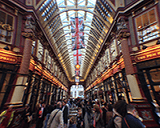}\vspace{2pt}
			\includegraphics[width=1\linewidth]{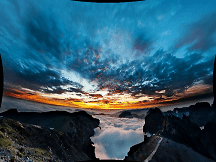}
	\end{minipage}}
	\hspace{-1.5ex}
	\subfigure[{Our method}]{
		\begin{minipage}[b]{0.19\linewidth}
			\includegraphics[width=1\linewidth]{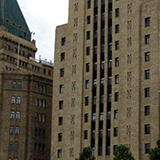}\vspace{2pt}
			\includegraphics[width=1\linewidth]{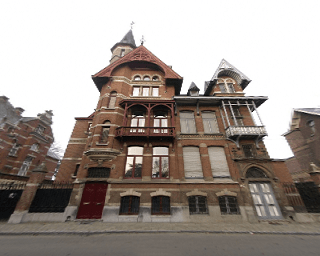}\vspace{2pt}
			\includegraphics[width=1\linewidth]{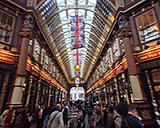}\vspace{2pt}
			\includegraphics[width=1\linewidth]{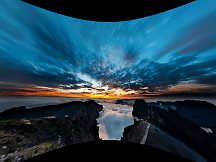}
	\end{minipage}}
\caption{Qualitative rectification comparison results on wild fisheye images with different distorted effects mined from the Internet. From left to right are the input fisheye images, rectification results of three state-of-the-art methods (Bukhari~\cite{bukhari2013automatic}, AlemnFlores~\cite{aleman2014automatic}, Rong~\cite{rong2016radial}), and our results.}
\label{fig:real2}
\end{figure*}

For the model trained only on $\text{SLF}_{\text{sun}}$ collection, the performance is not accurate since the training images are taken from the virtual 3D environments. By contrast, the collection $\text{SLF}_{\text{wf}}$ can improve the performance by a significant margin since the sourced images are taken from real-world environments. By utilizing all the training samples of the SLF dataset, we achieve the best performance.

\subsubsection{Results on Internet Images}
Considering that the images in~\cite{fishdataset2016-5} are still limited, we also fetch fisheye images with different types of distortion from the Internet for comparison. As shown in Fig.~{\ref{fig:real2}}, our proposed method has excellent rectification performance even for real fisheye images fetched from the Internet, which verifies that our network has strong rectification ability and has the potential to be used for uncontrolled environments without tuning.% generalization performance.

\subsection{Ablation Study}
\begin{figure*}[thp!]
    \centering
    \subfigure[{Fisheye}]{
    \begin{minipage}[b]{0.12\linewidth}
    \includegraphics[width=1\linewidth]{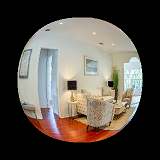}\vspace{1pt}
    \includegraphics[width=1\linewidth]{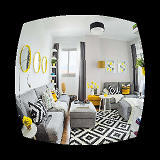}\vspace{1pt}
    \includegraphics[width=1\linewidth]{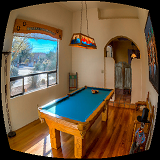}\vspace{1pt}
    \includegraphics[width=1\linewidth]{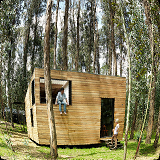}
    \end{minipage}}
    \hspace{-1.6ex}
    \subfigure[{$\text{BL}_\text{1}$}\label{fig:ablation-1}]{
    \begin{minipage}[b]{0.12\linewidth}
    \includegraphics[width=1\linewidth]{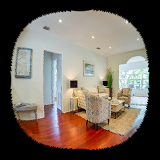}\vspace{1pt}
    \includegraphics[width=1\linewidth]{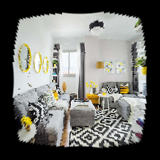}\vspace{1pt}
    \includegraphics[width=1\linewidth]{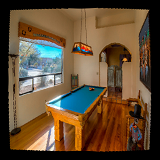}\vspace{1pt}
    \includegraphics[width=1\linewidth]{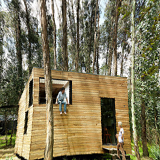}
    \end{minipage}}
    \hspace{-1.6ex}
    \subfigure[{$\text{BL}_\text{2}$}\label{fig:ablation-2}]{
    \begin{minipage}[b]{0.12\linewidth}
    \includegraphics[width=1\linewidth]{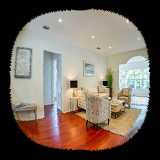}\vspace{1pt}
    \includegraphics[width=1\linewidth]{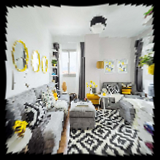}\vspace{1pt}
    \includegraphics[width=1\linewidth]{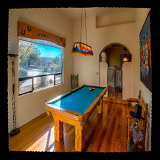}\vspace{1pt}
    \includegraphics[width=1\linewidth]{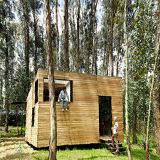}
    \end{minipage}}
    \hspace{-1.6ex}
    \subfigure[{$\text{BL}_\text{3}$}\label{fig:ablation-3}]{
    \begin{minipage}[b]{0.12\linewidth}
    \includegraphics[width=1\linewidth]{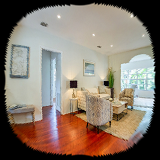}\vspace{1pt}
    \includegraphics[width=1\linewidth]{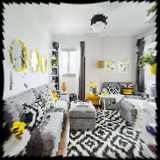}\vspace{1pt}
    \includegraphics[width=1\linewidth]{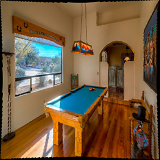}\vspace{1pt}
    \includegraphics[width=1\linewidth]{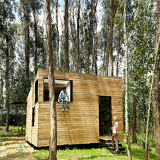}
    \end{minipage}}
    \hspace{-1.6ex}
    \subfigure[{$\text{BL}_\text{4}$}\label{fig:ablation-4}]{
    \begin{minipage}[b]{0.12\linewidth}
    \includegraphics[width=1\linewidth]{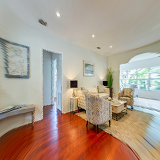}\vspace{1pt}
    \includegraphics[width=1\linewidth]{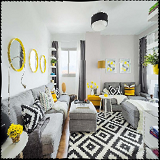}\vspace{1pt}
    \includegraphics[width=1\linewidth]{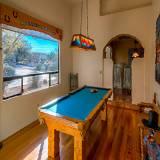}\vspace{1pt}
    \includegraphics[width=1\linewidth]{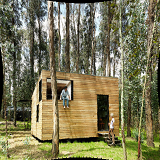}
    \end{minipage}}
     \hspace{-1.6ex}
    \subfigure[{$\text{BL}_\text{5}$}\label{fig:ablation-5}]{
    \begin{minipage}[b]{0.12\linewidth}
    \includegraphics[width=1\linewidth]{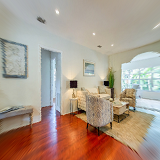}\vspace{1pt}
    \includegraphics[width=1\linewidth]{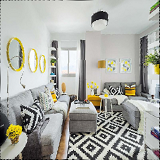}\vspace{1pt}
    \includegraphics[width=1\linewidth]{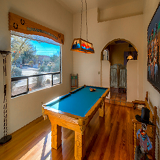}\vspace{1pt}
    \includegraphics[width=1\linewidth]{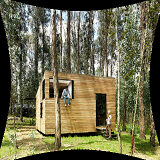}
    \end{minipage}}
     \hspace{-1.6ex}
    \subfigure[{$\text{BL}_\text{6}$}\label{fig:ablation-6}]{
    \begin{minipage}[b]{0.12\linewidth}
    \includegraphics[width=1\linewidth]{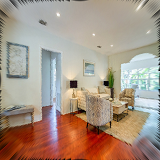}\vspace{1pt}
    \includegraphics[width=1\linewidth]{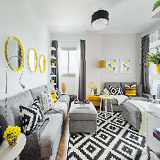}\vspace{1pt}
    \includegraphics[width=1\linewidth]{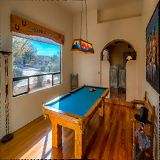}\vspace{1pt}
    \includegraphics[width=1\linewidth]{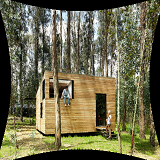}
    \end{minipage}}
     \hspace{-1.6ex}
    \subfigure[GT]{
    \begin{minipage}[b]{0.12\linewidth}
    \includegraphics[width=1\linewidth]{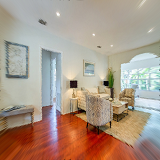}\vspace{1pt}
    \includegraphics[width=1\linewidth]{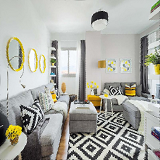}\vspace{1pt}
    \includegraphics[width=1\linewidth]{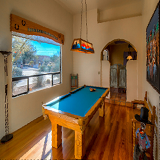}\vspace{1pt}
    \includegraphics[width=1\linewidth]{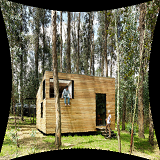}
    \end{minipage}}
    \vspace{-3mm}
    \caption{Visualization of ablation experiments; the $\text{BL}_\text{1}-\text{BL}_\text{6}$ represent different baselines, as shown in Tab.~\ref{tab:ablation-study}.}
    \label{fig:Ablation}
\end{figure*}

We conduct ablation studies on the proposed SLF dataset to justify the design decisions of our proposed network.  
First, we verify the effectiveness of using the learned heatmaps of distorted lines according to three ablation choices: $\text{BL}_{1}$: learning the distortion parameters only from the colored images; $\text{BL}_{2}$: learning the distortion parameters only from the learned heatmaps; and $\text{BL}_{3}$: learning the distortion parameters from the learned line geometry and the image appearance. In these experiments, we use only the global branch in the multiscale calibration module and remove the loss functions of $\mathcal{L}_{\textrm{geo}}$ and $\mathcal{L}_{\textrm{pix}}$.
As reported in Tab.~\ref{tab:ablation-study} and Fig.~\ref{fig:ablation-1}-\ref{fig:ablation-3}, the learned heatmaps of distorted lines are positive for fisheye image rectification. Compared with the choice of only using the colored images, the learned heatmaps of lines can obtain better results. Furthermore, the image appearance and the learned heatmaps are complementary to obtain the best performance.  

\begin{table}[t]
    \centering
    \caption{Ablation study of the proposed method on SLF dataset. $\text{BL}_{1}-\text{BL}_{6}$ represent different training strategies respectively.}
    \resizebox{\linewidth}{!}
{
    \begin{tabular}{c|cc|c|cc|ccc}\toprule
     strategy & \multicolumn{2}{c|}{Input} & \multirow{2}{*}{\specialcell{Multi-scale\\Calibration}} & \multicolumn{2}{c|}{Loss Functions} & \multicolumn{3}{c}{Metrics}\\
     & $I_f$ & $\hat{L}_f$ &  & $\mathcal{L}_{geo}$ & $\mathcal{L}_{pix}$ & PSNR & SSIM &RPE\\\hline
    $\text{BL}_\text{1}$ & \checkmark &  &   &  & & 16.24 & 0.61 & 4.51  \\
    $\text{BL}_\text{2}$ & & \checkmark & & & & 18.61 & 0.62 & 2.08 \\
    $\text{BL}_\text{3}$ & \checkmark & \checkmark & & & & 21.29 & 0.68 & 1.35 \\\midrule
    $\text{BL}_\text{4}$ & \checkmark & \checkmark & \checkmark & & & 25.65 & 0.80 & 0.91 \\
    $\text{BL}_\text{5}$ & \checkmark & \checkmark & \checkmark & \checkmark & & 27.83 & 0.88 & 0.43\\
    $\text{BL}_\text{6}$ & \checkmark & \checkmark & \checkmark & \checkmark & \checkmark & 28.46 & 0.90 &0.33\\\bottomrule
\end{tabular}
}
\label{tab:ablation-study}
\end{table}

Next, we verify the effectiveness of the multiscale calibration module and the additional loss functions of geometric constraints and estimation uncertainty. As reported in the third and fourth rows of Tab.~\ref{tab:ablation-study}, the multiscale calibration module further improves the performance. Compared with the results shown in  Fig.~\ref{fig:ablation-3}, the multiscale calibration module handles the nonlinear distortion better and improves the rectification results in Fig.~\ref{fig:ablation-4}. 

Finally, the loss functions of the geometric constraints and the uncertainty are verified. As reported in Tab.~\ref{tab:ablation-study}, the loss function $\mathcal{L}_{\textrm{geo}}$ significantly improves the metrics of SSIM and RPE, which demonstrates that the geometric constraint is positive for improving the structural accuracy of the rectification results. The images shown in Fig.~\ref{fig:ablation-4} and Fig.~\ref{fig:ablation-5} qualitatively verify this improvement. When the loss of estimation uncertainty is added to the training strategy, LaRecNet achieves the best rectification performance.

\section{Conclusion}\label{sec:conclusion}
In this paper, we propose a novel and efficient model named LaRecNet that utilizes line constraints to calibrate fisheye lenses and eliminate distortion parameters automatically from only a single image. Specifically, we adopt line-aware strategies and a multiscale calibration module to learn how to straighten the distorted lines, and an attentive regularization module is proposed to obtain uncertainty maps that provide the uncertainty-guided pixel loss to solve the issue of inaccurate distorted line labels. Extensive experimental results on synthesized datasets as well as real fisheye images demonstrate that our method performs much better than present state-of-the-art methods. To better train the proposed network, we also reuse the existing datasets that have rich 2D and 3D geometric information to generate a synthetic dataset for fisheye calibration.

Since the geometric constraints of a single image are limited and the data distribution of the synthesized dataset is different from that of the real world, some real fisheye images cannot be well rectified. Therefore, we plan to focus on image processing based on videos or image sequences to obtain a better rectification effect in the future while extending our dataset by selecting fisheye images in the real world.

\bibliographystyle{IEEEtran}
\bibliography{egbib}

\begin{thebibliography}{10}
\providecommand{\url}[1]{#1}
\csname url@samestyle\endcsname
\providecommand{\newblock}{\relax}
\providecommand{\bibinfo}[2]{#2}
\providecommand{\BIBentrySTDinterwordspacing}{\spaceskip=0pt\relax}
\providecommand{\BIBentryALTinterwordstretchfactor}{4}
\providecommand{\BIBentryALTinterwordspacing}{\spaceskip=\fontdimen2\font plus
\BIBentryALTinterwordstretchfactor\fontdimen3\font minus
  \fontdimen4\font\relax}
\providecommand{\BIBforeignlanguage}[2]{{%
\expandafter\ifx\csname l@#1\endcsname\relax
\typeout{** WARNING: IEEEtran.bst: No hyphenation pattern has been}%
\typeout{** loaded for the language `#1'. Using the pattern for}%
\typeout{** the default language instead.}%
\else
\language=\csname l@#1\endcsname
\fi
#2}}
\providecommand{\BIBdecl}{\relax}
\BIBdecl

\bibitem{weng1992camera}
J.~Weng, P.~Cohen, and M.~Herniou, ``Camera calibration with distortion models
  and accuracy evaluation,'' \emph{IEEE TPAMI}, vol.~14, no.~10, pp. 965--980,
  1992.

\bibitem{ricolfe2010lens}
C.~Ricolfe-Viala and A.-J. Sanchez-Salmeron, ``Lens distortion models
  evaluation,'' \emph{Applied optics}, vol.~49, no.~30, pp. 5914--5928, 2010.

\bibitem{yang2005nonlinear}
Y.~Yang, J.~X. Chen, and M.~Beheshti, ``Nonlinear perspective projections and
  magic lenses: 3d view deformation,'' \emph{IEEE Computer Graphics and
  Applications}, vol.~25, no.~1, pp. 76--84, 2005.

\bibitem{bertozzi2000vision}
M.~Bertozzi, A.~Broggi, and A.~Fascioli, ``Vision-based intelligent vehicles:
  State of the art and perspectives,'' \emph{ROBOT AUTON SYST}, vol.~32, no.~1,
  pp. 1--16, 2000.

\bibitem{xiong1997creating}
Y.~Xiong and K.~Turkowski, ``Creating image-based vr using a self-calibrating
  fisheye lens,'' in \emph{CVPR}, 1997.

\bibitem{huang20176}
J.~Huang, Z.~Chen, D.~Ceylan, and H.~Jin, ``6-dof vr videos with a single
  360-camera,'' in \emph{VIRTUAL REAL-LONDON}, 2017.

\bibitem{szeliski1997creating}
R.~Szeliski and H.-Y. Shum, ``Creating full view panoramic image mosaics and
  environment maps,'' in \emph{SIGGRAPH}, 1997.

\bibitem{zhang2000flexible}
Z.~Zhang, ``A flexible new technique for camera calibration,'' \emph{IEEE
  TPAMI}, vol.~22, no.~11, pp. 1330--1334, 2000.

\bibitem{Grossberg2001A}
M.~D. Grossberg and S.~K. Nayar, ``A general imaging model and a method for
  finding its parameters,'' in \emph{ICCV}, 2001.

\bibitem{Sturm2004A}
P.~Sturm and S.~Ramalingam, ``A generic concept for camera calibration,'' in
  \emph{ECCV}, 2004.

\bibitem{kannala2006generic}
J.~Kannala and S.~S. Brandt, ``A generic camera model and calibration method
  for conventional, wide-angle, and fish-eye lenses,'' \emph{IEEE TPAMI},
  vol.~28, no.~8, pp. 1335--1340, 2006.

\bibitem{Scaramuzza2006A}
D.~Scaramuzza, A.~Martinelli, and R.~Siegwart, ``A flexible technique for
  accurate omnidirectional camera calibration and structure from motion,'' in
  \emph{ICVS}, 2006.

\bibitem{heikkila2000geometric}
J.~Heikkila, ``Geometric camera calibration using circular control points,''
  \emph{IEEE TPAMI}, vol.~22, no.~10, pp. 1066--1077, 2000.

\bibitem{stein1997lens}
G.~P. Stein, ``Lens distortion calibration using point correspondences,'' in
  \emph{CVPR}, 1997.

\bibitem{Faugeras1992Camera}
O.~Faugeras, Q.~Luong, and S.~Maybank, ``Camera self-calibration: Theory and
  experiments,'' in \emph{ECCV}, 1992.

\bibitem{Maybank1992A}
S.~J. Maybank and O.~D. Faugeras, ``A theory of self-calibration of a moving
  camera,'' \emph{IJCV}, vol.~8, pp. 123--151, 1992.

\bibitem{devernay2001straight}
F.~Devernay and O.~Faugeras, ``Straight lines have to be straight,'' \emph{MACH
  VISION APPL}, vol.~13, no.~1, pp. 14--24, 2001.

\bibitem{Range}
D.~C. Brown, ``Close-range camera calibration,'' \emph{PHOTOGRAMMETRIC ENG},
  vol.~37, no.~8, pp. 855--866, 1971.

\bibitem{barreto2009automatic}
J.~Barreto, J.~Roquette, P.~Sturm, and F.~Fonseca, ``Automatic camera
  calibration applied to medical endoscopy,'' in \emph{BMVC}, 2009.

\bibitem{melo2013unsupervised}
R.~Melo, M.~Antunes, J.~P. Barreto, G.~Falcao, and N.~Goncalves, ``Unsupervised
  intrinsic calibration from a single frame using a "plumb-line" approach,'' in
  \emph{ICCV}, 2013.

\bibitem{bukhari2013automatic}
F.~Bukhari and M.~N. Dailey, ``Automatic radial distortion estimation from a
  single image,'' \emph{J MATH IMAGING VIS}, vol.~45, no.~1, pp. 31--45, 2013.

\bibitem{aleman2014automatic}
M.~Alem{\'a}n-Flores, L.~Alvarez, L.~Gomez, and D.~Santana-Cedr{\'e}s,
  ``Automatic lens distortion correction using one-parameter division models,''
  \emph{IPOL}, vol.~4, pp. 327--343, 2014.

\bibitem{zhang2015line}
M.~Zhang, J.~Yao, M.~Xia, K.~Li, Y.~Zhang, and Y.~Liu, ``Line-based multi-label
  energy optimization for fisheye image rectification and calibration,'' in
  \emph{CVPR}, 2015.

\bibitem{rong2016radial}
J.~Rong, S.~Huang, Z.~Shang, and X.~Ying, ``Radial lens distortion correction
  using convolutional neural networks trained with synthesized images,'' in
  \emph{ACCV}, 2016.

\bibitem{yin2018fisheyerecnet}
X.~Yin, X.~Wang, J.~Yu, M.~Zhang, P.~Fua, and D.~Tao, ``Fisheyerecnet: A
  multi-context collaborative deep network for fisheye image rectification,''
  in \emph{ECCV}, 2018.

\bibitem{XieT17}
S.~Xie and Z.~Tu, ``Holistically-nested edge detection,'' \emph{IJCV}, vol.
  125, no. 1-3, pp. 3--18, 2017.

\bibitem{ManinisPAG18}
K.~Maninis, J.~Pont{-}Tuset, P.~Arbelaez, and L.~V. Gool, ``Convolutional
  oriented boundaries: From image segmentation to high-level tasks,''
  \emph{IEEE TPAMI}, vol.~40, no.~4, pp. 819--833, 2018.

\bibitem{LiuCHBZBT19}
Y.~Liu, M.~Cheng, X.~Hu, J.~Bian, L.~Zhang, X.~Bai, and J.~Tang, ``Richer
  convolutional features for edge detection,'' \emph{IEEE TPAMI}, vol.~41,
  no.~8, pp. 1939--1946, 2019.

\bibitem{huang2018learning}
K.~Huang, Y.~Wang, Z.~Zhou, T.~Ding, S.~Gao, and Y.~Ma, ``Learning to parse
  wireframes in images of man-made environments,'' in \emph{CVPR}, 2018.

\bibitem{xue-lsd-pami}
N.~Xue, S.~Bai, F.-D. Wang, G.-S. Xia, T.~Wu, L.~Zhang, and P.~H. Torr,
  ``Learning regional attraction for line segment detection,'' \emph{IEEE
  TPAMI}, pp. 1--1, 2019.

\bibitem{Xue-HAWP20}
N.~Xue, T.~Wu, S.~Bai, F.-D. Wang, G.-S. Xia, L.~Zhang, and P.~H. Torr,
  ``Holistically-attracted wireframe parsing,'' in \emph{CVPR}, 2020.

\bibitem{song2016ssc}
S.~Song, F.~Yu, A.~Zeng, A.~X. Chang, M.~Savva, and T.~Funkhouser, ``Semantic
  scene completion from a single depth image,'' in \emph{CVPR}, 2017.

\bibitem{xue2019learning}
Z.-C. Xue, N.~Xue, G.-S. Xia, and W.~Shen, ``Learning to calibrate straight
  lines for fisheye image rectification,'' in \emph{CVPR}, 2019.

\bibitem{snyder1997flattening}
J.~P. Snyder, \emph{Flattening the earth: two thousand years of map
  projections}.\hskip 1em plus 0.5em minus 0.4em\relax University of Chicago
  Press, 1997.

\bibitem{miyamoto1964fish}
K.~Miyamoto, ``Fish eye lens,'' \emph{JOSA}, vol.~54, no.~8, pp. 1060--1061,
  1964.

\bibitem{TardifSTR09}
J.~Tardif, P.~F. Sturm, M.~Trudeau, and S.~Roy, ``Calibration of cameras with
  radially symmetric distortion,'' \emph{IEEE TPAMI}, vol.~31, no.~9, pp.
  1552--1566, 2009.

\bibitem{barreto2005geometric}
J.~P. Barreto and H.~Araujo, ``Geometric properties of central catadioptric
  line images and their application in calibration,'' \emph{IEEE TPAMI},
  vol.~27, no.~8, pp. 1327--1333, 2005.

\bibitem{Han_2017_CVPR}
D.~Han, J.~Kim, and J.~Kim, ``Deep pyramidal residual networks,'' in
  \emph{CVPR}, 2017.

\bibitem{Newell2016Stacked}
A.~Newell, K.~Yang, and J.~Deng, ``Stacked hourglass networks for human pose
  estimation,'' in \emph{ECCV}, 2016.

\bibitem{he2016deep}
K.~He, X.~Zhang, S.~Ren, and J.~Sun, ``Deep residual learning for image
  recognition,'' in \emph{CVPR}, 2016.

\bibitem{kendall2018multi}
A.~Kendall, Y.~Gal, and R.~Cipolla, ``Multi-task learning using uncertainty to
  weigh losses for scene geometry and semantics,'' in \emph{CVPR}, 2018.

\bibitem{blender2014blender}
{Blender Online Community}, ``Blender - a 3d modelling and rendering package,''
  Blender Foundation, Blender Institute Amsterdam, 2014.

\bibitem{PyTorch}
A.~Paszke, S.~Gross, F.~Massa, A.~Lerer, J.~Bradbury, G.~Chanan, T.~Killeen,
  and Z.~Lin, ``Pytorch: An imperative style, high-performance deep learning
  library,'' in \emph{NeuraIPS}, 2019.

\bibitem{fishdataset2016-5}
A.~Eichenseer and A.~Kaup, ``A data set providing synthetic and real-world
  fisheye video sequences,'' in \emph{ICASSP}, 2016.

\bibitem{SSIM}
Z.~Wang, A.~C. Bovik, H.~R. Sheikh, and E.~P. Simoncelli, ``Image quality
  assessment: from error visibility to structural similarity,'' \emph{IEEE
  TIP}, vol.~13, no.~4, pp. 600--612, 2004.

\bibitem{Martin2004Learning}
D.~R. Martin, C.~C. Fowlkes, and J.~Malik, ``Learning to detect natural image
  boundaries using local brightness, color, and texture cues,'' \emph{IEEE
  TPAMI}, vol.~26, no.~5, pp. 530--549, 2004.

\bibitem{scaramuzza2006toolbox}
D.~Scaramuzza, A.~Martinelli, and R.~Siegwart, ``A toolbox for easily
  calibrating omnidirectional cameras,'' in \emph{IROS}, 2006.

\end{thebibliography}

\end{document}